\documentclass[letterpaper, 10 pt, conference]{ieeeconf}

\IEEEoverridecommandlockouts
\usepackage{placeins}
\usepackage{amssymb}
\usepackage{multirow}
\usepackage{amsmath}
\usepackage{subcaption}
\usepackage{graphicx}
\usepackage{url}
\usepackage{booktabs}
\usepackage{tabularx}
\usepackage[table]{xcolor}
\usepackage{array}
\usepackage{xcolor}
\usepackage{dsfont}
\usepackage{graphics}
\usepackage{epsfig}
\usepackage{mathptmx}
\usepackage{times}
\usepackage{amsmath}
\usepackage{amssymb}
\usepackage{booktabs}
\usepackage{bm}
\usepackage{cuted}
\usepackage{dblfloatfix}
\begin{document}
\title{\LARGE \bf
UniReLo: Learning a Unified Humanoid Policy from Fall Recovery to
Locomotion across Diverse Terrains
}

\author{%
Xiaoyu Xu, Zhiming Chen, Yuenan Zhao, Xiang Zhang, Ran Song, and
Wei Zhang$^{*}$%
\thanks{All authors are with the School of Control Science and Engineering,
Shandong University, Jinan 250061, China. Xiaoyu Xu, Yuenan Zhao, Ran Song,
and Wei Zhang are also with the Key Laboratory of Machine Intelligence and
System Control, Ministry of Education, China.}%
\thanks{$^{*}$Corresponding author: Wei Zhang
(\texttt{davidzhang@sdu.edu.cn}).}%
}

\maketitle

\thispagestyle{empty}
\pagestyle{empty}


\begin{strip}
    \centering
    \includegraphics[width=\textwidth]{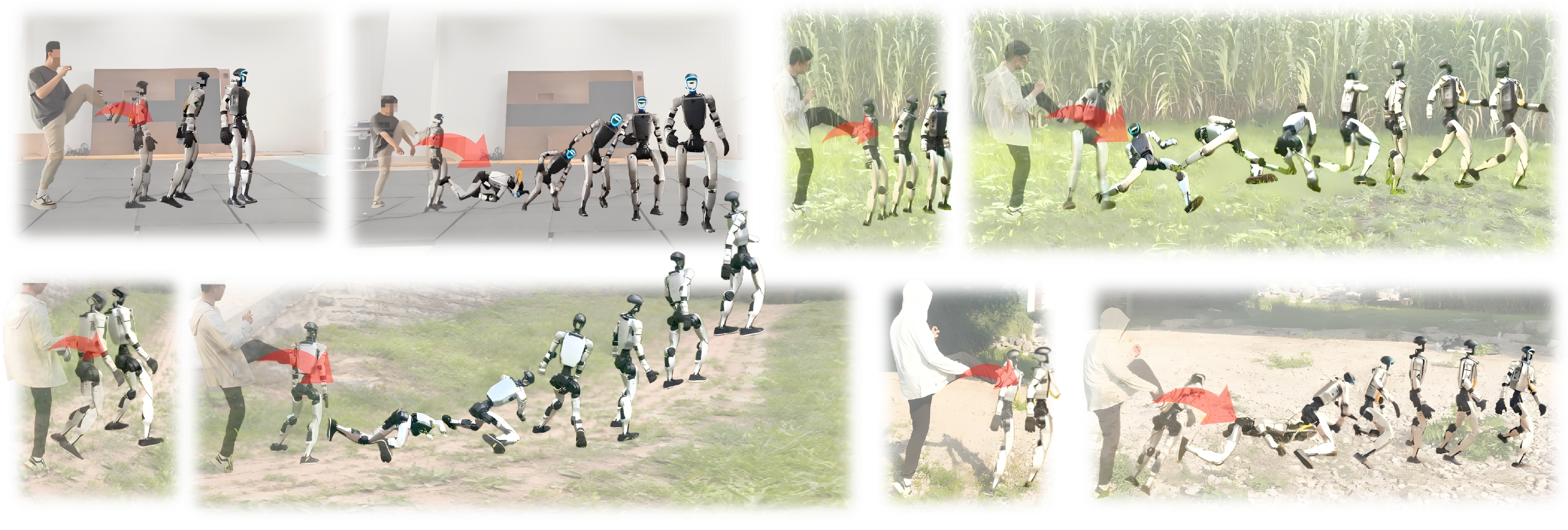}
    \captionof{figure}{Real-world disturbance-rejection and recovery-to-locomotion behaviors ofUniReLo across indoor flat ground and representative outdoor field terrains. Under moderate perturbations, the robot maintains locomotion through coordinated whole-body regulation and adaptive foot placement. Under fall-inducing perturbations, it performs whole-body recovery, re-establishes terrain-compatible support, and resumes commanded locomotion. Red arrows denote external disturbances.
    }
    \label{fig:teaser}
    \vspace{2pt}
\end{strip}

\begin{abstract}
Reliable fall recovery, which commonly aims at attaining a nominal upright posture, is essential for the autonomous operation of humanoid robots in unstructured field environments. Although existing posture-centered methods can synthesize coordinated whole-body recovery motions from diverse fallen configurations, they may result in a dynamically fragile support state, leading to secondary loss of balance or unstable resumption of commanded locomotion, particularly under terrain-dependent contact conditions. We propose to learn a unified humanoid policy from fall recovery to locomotion (UniReLo) across heterogeneous field terrains. UniReLo leverages continuously gated multi-scale motion priors to modulate frame-, sequence-, and gait-level adversarial supervision according to recovery progress, preserving the distinct temporal structures of recovery and locomotion without requiring fixed-threshold switching. In addition, terrain-conditioned recovery guidance evaluates the evolving
support state using a terrain-relative support representation and
support-feasibility assessment. Simulation and outdoor real-world experiments demonstrate that UniReLo can deliver stable and continuous recovery-to-locomotion behaviors for humanoids across diverse field terrains. The supplementary video is available at
\url{https://vsislab.github.io/UniReLo/}.
\end{abstract}

\section{INTRODUCTION}
\label{sec:introduction}
Humanoid robots operating in unstructured field environments are vulnerable
to falls caused by uneven terrain, uncertain contacts, surface deformation,
and external disturbances. Autonomous fall recovery is therefore critical
to limiting task interruption, avoiding secondary falls, and sustaining
operation without human assistance. For humanoids, however, successful
recovery involves more than returning the body to an upright posture. The
contact configuration, whole-body posture, and momentum established during
the rising motion jointly determine whether the robot can regain dynamic
balance and continue task execution. This requirement becomes particularly
important in outdoor environments, where loose, inclined, and irregular
surfaces can make a posture that is recoverable on rigid flat ground
unsuitable for continued locomotion. Fall recovery should thus be viewed as
a continuous process of restoring dynamically viable task capability,
rather than as an isolated posture-reaching maneuver. 

Recent advances in reinforcement learning (RL) have enabled end-to-end humanoid
standing-up and fall recovery from diverse fallen configurations, without handcrafted motion scripts or prescribed contact sequences~\cite{huang2025host}. However, in the absence of reference-motion constraints, motion quality must be induced indirectly through task-oriented rewards and is therefore sensitive to reward shaping and exploration design. With the development of motion-prior-based RL, methods such as DeepMimic and Adversarial Motion Priors (AMP) showed that reference motions can regularize policy learning and improve the physical plausibility and style consistency of learned behaviors ~\cite{peng2018deepmimic,peng2021amp}. This provides a promising foundation for integrating multiple behaviors within a single policy. Nevertheless, a policy spanning nonperiodic whole-body recovery, transient support reorganization, and periodic locomotion must represent motion distributions with markedly different temporal structures. A shared, undifferentiated adversarial prior does not explicitly preserve these distinctions, whereas hard mode-dependent routing reintroduces discrete behavioral boundaries. Learning a unified policy therefore requires motion supervision that distinguishes heterogeneous temporal structures while adapting continuously to the evolving recovery state, enabling a smooth progression from whole-body recovery through support reorganization to resumed locomotion. A further challenge arises from the terrain dependence of recovery
viability. In field environments, successful rising cannot be determined by
body height or uprightness alone, because the resulting support state
may still exhibit insufficient support margin, a terrain-incompatible knee configuration, foot slip, or downhill drift. Such states may appear nominally recovered but remain unable to sustain commanded locomotion, leading to secondary loss of balance. Moreover, the recoverability of the same fallen pose may vary
substantially across rigid ground, loose gravel, and inclined terrain~\cite{he2025getup}.

This paper proposes UniReLo, a framework for learning a unified humanoid policy from fall recovery to velocity-commanded locomotion across heterogeneous field terrains, as shown in Fig.~\ref{fig:teaser}. Rather than defining recovery by the attainment of a nominal upright posture, UniReLo guides the robot toward a terrain-compatible support state that can sustain continued command execution. In UniReLo, continuously gated multi-scale motion priors first modulate frame-, sequence-, and gait-level adversarial supervision according to recovery progress, enabling motion guidance to evolve smoothly across whole-body recovery, support reorganization, and commanded locomotion. Then, terrain-conditioned recovery guidance constructs a terrain-relative support representation and performs support-feasibility assessment to derive a physically grounded measure of recovery progress. In addition, terrain-pose plasticity-aware initialization is employed during training to increase exposure to terrain-sensitive recovery configurations.  All components are jointly optimized within a single proprioceptive policy and implemented on a 29-DoF Unitree G1 humanoid, without terrain-specific policy selection during deployment.

Our main contributions are summarized as follows:
\begin{itemize}

\item 
We formulate fall recovery as restoring a terrain-compatible, dynamically viable state for continued command execution, enabling a single policy to unify whole-body recovery, support reorganization, and velocity-commanded locomotion across diverse field terrains.

\item 
We modulate frame-, sequence-, and gait-level adversarial supervision
according to recovery progress, enabling motion guidance to evolve smoothly across whole-body recovery, support reorganization, and commanded locomotion, while preserving their distinct temporal characteristics.

\item 
We develop terrain-conditioned recovery guidance that combines a
terrain-relative support representation with support-feasibility assessment,
providing a continuous measure of progress toward terrain-compatible,
locomotion-ready states.
\end{itemize}

\begin{figure*}[t]
    \centering
    \includegraphics[width=\linewidth]{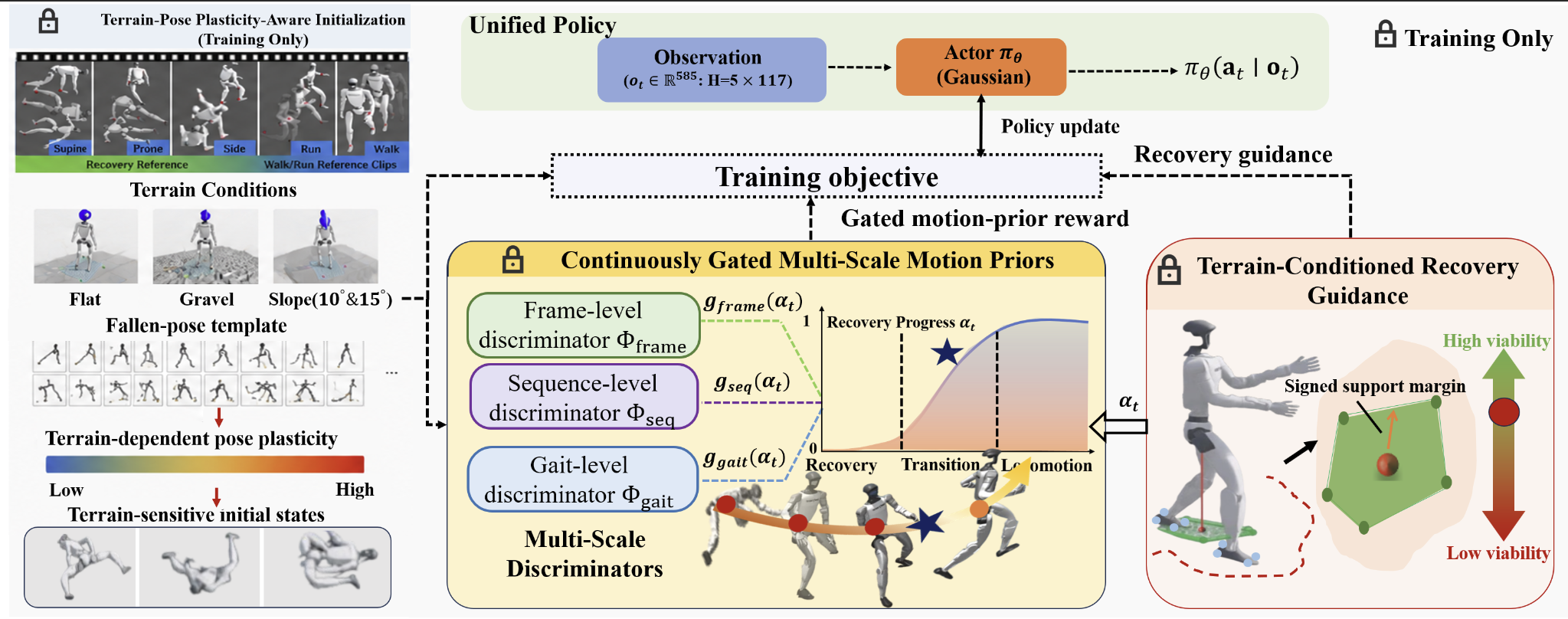}
    \caption{Overview of the proposed UniReLo framework. Terrain-conditioned recovery guidance constructs a terrain-relative support representation, evaluates support feasibility, and produces recovery progress $\alpha_t$, which is used by the continuously gated multi-scale motion priors to coordinate
frame-, sequence-, and gait-level adversarial supervision. Terrain-pose
plasticity-aware initialization complements these mechanisms by emphasizing terrain-sensitive fallen configurations during training. All training signals jointly update the unified policy, while only the proprioceptive actor is retained for deployment.}
    \label{fig:fig2}
\end{figure*}

\section{RELATED WORK}
\label{sec:related}
This section reviews related work in two areas: learning-based humanoid
recovery and motion priors, and terrain-adaptive recovery-to-locomotion control.
\subsection{Learning-Based Humanoid Recovery and Motion Priors}
\label{subsec:recovery_learning}

Humanoid fall recovery has traditionally been addressed through contact
planning, trajectory optimization, and model-based balance control
~\cite{mordatch2012contact,carius2021trajectory,pratt2006capture}.
Although these approaches provide interpretable contact sequences and
dynamically consistent trajectories, they often rely on predefined contact models, recovery templates, or online optimization. Recent deep
reinforcement learning methods instead learn feedback policies directly from diverse fallen configurations, reducing the need for handcrafted motion scripts and prescribed contact sequences ~\cite{gaspard2024frasa,chen2025hifar,he2025getup,
huang2025host}. Related work further couples fall mitigation with post-fall recovery~\cite{xu2025firm}. For methods trained primarily through
task-oriented rewards, however, motion smoothness, physical plausibility, and temporal coherence must be induced indirectly through reward shaping, exploration design, and additional regularization, which is particularly challenging for nonperiodic and contact-rich whole-body recovery.

Reference-motion-guided reinforcement learning provides a complementary
means of shaping recovery behaviors. DeepMimic~\cite{peng2018deepmimic}
tracks structured motion references, whereas AMP~\cite{peng2021amp} learns
data-driven style rewards from reference-motion distributions. Existing
multi-skill AMP variants commonly associate different reference datasets or
motion priors with predefined skill commands or heuristic state partitions.
Multiple AMP enables discrete switching among motion styles through explicit
skill conditioning~\cite{vollenweider2023mamp}. For recovery-to-locomotion control, Lu et al.~\cite{lu2026stateamp} divide recovery and locomotion into two discrete behavioral stages and assign the corresponding motion samples according to fixed state thresholds, making motion-prior selection dependent on predefined transition criteria.

\subsection{Terrain-Adaptive and Recovery-to-Locomotion Control}
\label{subsec:terrain_unified}

Terrain geometry, friction, compliance, and contact uncertainty substantially
affect foothold feasibility, support stability, and whole-body posture.
Terrain-adaptive locomotion policies address these variations through domain
randomization, terrain curricula, observation histories, implicit terrain
estimation, and exteroceptive
perception~\cite{lee2020learning,rudin2022learning,
nahrendra2023dreamwaq,miki2022learning,gu2024dwl,
radosavovic2024learning,sun2025wmr,sun2025perceptive}.
Beyond general terrain adaptation, recent studies show that terrain
constraints must be considered together with the contact requirements and
whole-body dynamics of the target task. For locomotion on sparse feasible
contact regions, Walk the  PLANC~\cite{dai2026planc} coordinates balance,
foothold placement, and step timing on constrained terrains such as stepping
stones, beams, and planks, where small contact or timing errors may cause
irrecoverable failure. 

For fall recovery, FR-Net~\cite{lu2025frnet}
addresses incomplete terrain perception and uncertain interactions on
challenging surfaces through explicit mass-contact prediction, enabling
quadrupedal recovery across diverse terrains, including steep stairs.
Terrain interaction is equally critical for humanoid standing-up and
fall-recovery control. HumanUP~\cite{he2025getup} extends humanoid
getting-up to deformable, slippery, and inclined terrains. Its experiments
show that the robot may partially rise on grass slopes and snow-covered
surfaces, but subsequently fall because of unstable foot placement or
surface slippage. HoST~\cite{huang2025host} learns standing-up motions from
diverse terrain-supported postures, and its cross-terrain analysis reports
higher balancing-energy consumption on sloped terrain.
FIRM~\cite{xu2025firm} further unifies fall mitigation and recovery within
a single policy and evaluates recovery on flat, uneven, wave, and rough
terrains, with recovery success varying substantially across terrain
conditions.

\section{METHOD}
\label{sec:method}

This section presents the UniReLo framework, as illustrated in
Fig.~\ref{fig:fig2}. We first formulate the unified recovery-to-locomotion control problem and introduce continuously gated multi-scale motion priors for coordinating recovery, support reorganization, and locomotion.
We then construct terrain-conditioned recovery guidance to evaluate support viability and derive the continuous recovery progress, followed
by terrain-pose plasticity-aware initialization for training-state sampling.

\subsection{Problem Formulation and Objective}
\label{subsec:overview}

A single policy
$\pi_\theta(\mathbf a_t\mid\mathbf o_t)$
maps proprioceptive observations to joint-position targets tracked by a
low-level controller. The observation $\mathbf o_t$ contains joint positions
$\mathbf q_t$, joint velocities $\dot{\mathbf q}_t$, projected gravity
$\mathbf g_t^b$, base angular velocity $\boldsymbol{\omega}_t$, previous
actions, and the velocity command
$\mathbf u_t=
[v_{x,t}^{\mathrm{cmd}},
v_{y,t}^{\mathrm{cmd}},
\omega_{z,t}^{\mathrm{cmd}}]^\top$.

Training covers
$\mathcal K=
\{\mathrm{flat},\mathrm{gravel},
\mathrm{slope}\text{-}10^\circ,
\mathrm{slope}\text{-}15^\circ\}$.
We distinguish the policy observation $\mathbf o_t$, the privileged
simulator state $\mathbf s_t$ used only for constructing training signals,
and the temporal motion windows evaluated by the motion-prior branches.

The policy is optimized with the total reward

\begin{equation}
R_t
=
R_t^{\mathrm{task}}
+
R_t^{\mathrm{TCG}}
+
\rho M_t^{\mathrm{style}}
-
R_t^{\mathrm{reg}},
\label{eq:total_reward}
\end{equation}

where $R_t^{\mathrm{task}}$ coordinates rising and command tracking,
$R_t^{\mathrm{TCG}}$ provides terrain-conditioned recovery guidance,
$M_t^{\mathrm{style}}$ denotes the gated motion-prior reward, and
$R_t^{\mathrm{reg}}$ collects nonnegative control and safety costs.
A continuous recovery-progress variable $\alpha_t\in[0,1]$ coordinates
both the task objectives and motion priors.

\subsection{Continuously Gated Multi-Scale Motion Priors}
\label{subsec:cg}

Whole-body recovery, transient support reorganization, and periodic
locomotion exhibit distinct temporal structures. A temporally
undifferentiated motion prior may therefore obscure their characteristic
motion patterns. UniReLo employs three specialized AMP discriminators:
a frame-level branch $\Phi_{\mathrm{frame}}$ operating on a single frame,
a sequence-level branch $\Phi_{\mathrm{seq}}$ operating on a five-frame
window, and a gait-level branch $\Phi_{\mathrm{gait}}$ operating on a
ten-frame window. The branches share the same architecture but have
independent parameters.

The frame branch evaluates instantaneous posture consistency, the sequence
branch captures short-horizon recovery and support-reorganization dynamics,
and the gait branch evaluates longer-horizon locomotion regularity.

\begin{figure}[!t]
\centering
\includegraphics[width=\linewidth]{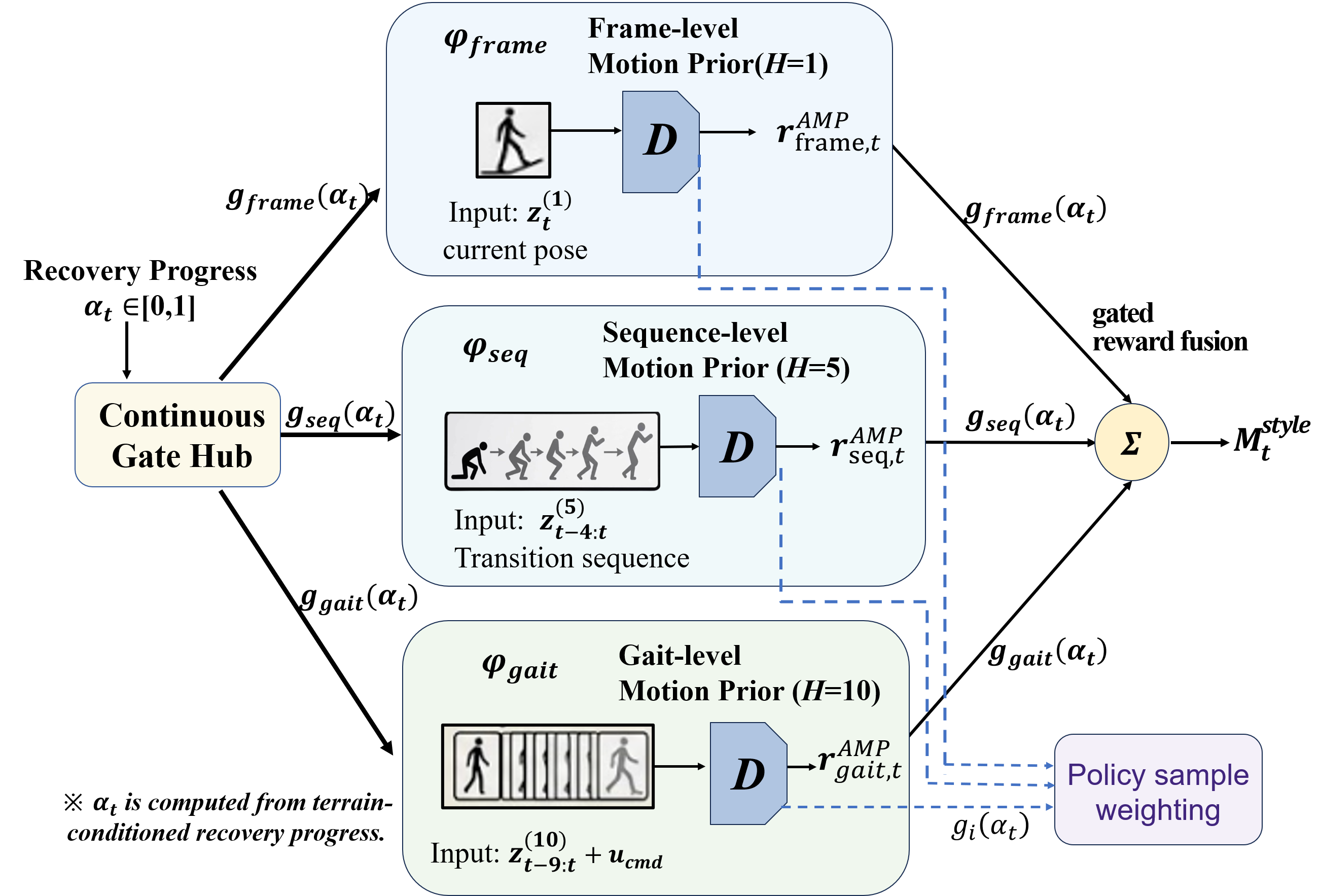}
\caption{
Continuously gated multi-scale motion priors. Recovery progress
$\alpha_t$ continuously adjusts the contributions of the frame-, sequence-,
and gait-level motion priors. The resulting gates regulate both motion-prior
reward fusion and policy-sample weighting during discriminator training.
}
\label{fig:fig3}
\end{figure}

Let
$\mathcal I=\{\mathrm{frame},\mathrm{seq},\mathrm{gait}\}$.
The normalized gate vector is

\begin{equation}
\mathbf g_t
=
\frac{1}{Z_t}
\begin{bmatrix}
(1-\alpha_t)^2\\
4\alpha_t(1-\alpha_t)\\
\alpha_t^2
\end{bmatrix},
\qquad
Z_t=1+2\alpha_t(1-\alpha_t),
\label{eq:gates}
\end{equation}

where
$\mathbf g_t=
[g_{\mathrm{frame}}(\alpha_t),
g_{\mathrm{seq}}(\alpha_t),
g_{\mathrm{gait}}(\alpha_t)]^\top$.
The gates remain nonnegative and satisfy
$\sum_{i\in\mathcal I}g_i(\alpha_t)=1$.

For branch $i$, the least-squares motion-prior reward is

\begin{equation}
r_{i,t}^{\mathrm{AMP}}
=
\max
\left[
0,\,
1-\frac{1}{4}
\left(
\Phi_i(\mathbf z_{t-H_i+1:t})-1
\right)^2
\right],
\label{eq:branch_amp_reward}
\end{equation}

where $H_i\in\{1,5,10\}$ is the temporal horizon of the corresponding
branch. The gated motion-prior reward is

\begin{equation}
M_t^{\mathrm{style}}
=
\sum_{i\in\mathcal I}
g_i(\alpha_t)
r_{i,t}^{\mathrm{AMP}}.
\label{eq:style_reward}
\end{equation}

Frame-level supervision is emphasized during early recovery,
sequence-level supervision becomes prominent during support reorganization,
and gait-level supervision gradually dominates after locomotion is
established. Their overlapping activation avoids fixed
recovery-to-locomotion thresholds.

The same gates also regulate discriminator specialization. For branch $i$,
the policy-generated component of the discriminator objective is

\begin{equation}
\mathcal L_i^{\pi}
=
\mathbb E_{(\mathbf z^\pi,\alpha)\sim\mathcal B_\pi}
\left[
g_i(\alpha)
\left(
\Phi_i(\mathbf z^\pi)+1
\right)^2
\right],
\qquad i\in\mathcal I .
\label{eq:gated_policy_samples}
\end{equation}

Here, $\mathcal B_\pi$ is the policy-motion buffer. Reference windows remain
unweighted and follow the standard least-squares AMP objective
\cite{peng2021amp}. Consequently, each discriminator focuses on policy
segments consistent with its temporal specialization, while the three gated
rewards jointly update the shared actor.

\subsection{Terrain-Conditioned Recovery Guidance}
\label{subsec:tcg}

\begin{figure}[!t]
\centering
\includegraphics[width=\linewidth]{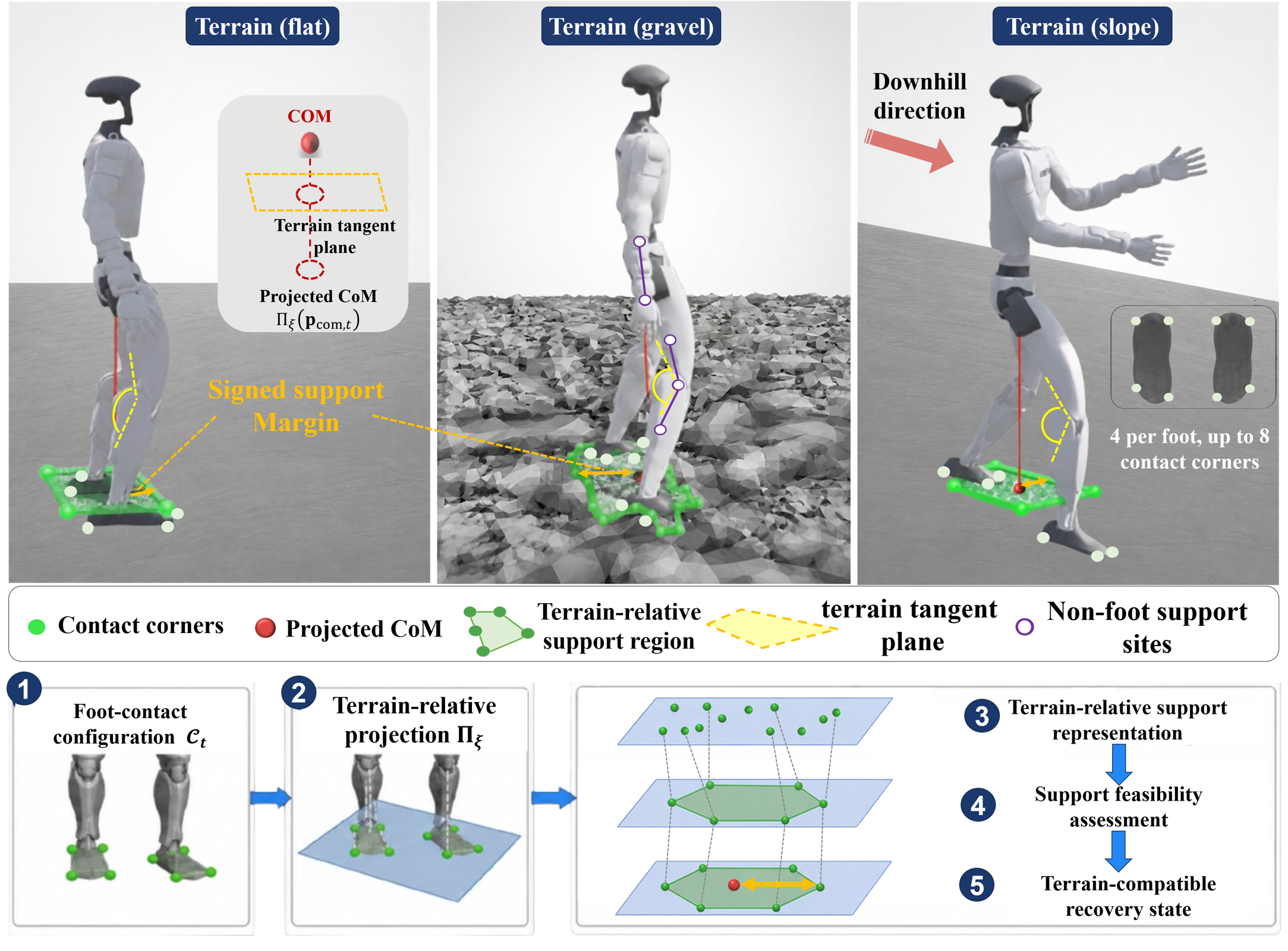}
\caption{
Terrain-conditioned recovery guidance. Active foot-contact corners
$\mathcal C_t$ are projected onto the local terrain tangent plane to
construct the support representation $S_{\xi,t}$. Green dots denote
foot-contact corners, purple hollow markers identify representative non-foot
support sites, and the red dot denotes the projected center of mass. Signed
support margin, contact redundancy, terrain-dependent knee configuration,
foot slip, and downhill drift jointly determine recovery viability and the
continuous progress variable $\alpha_t$.
}
\label{fig:fig4}
\end{figure}

Recovery progress should reflect compatibility with stable locomotion under
the current terrain rather than body elevation alone. As illustrated in
Fig.~\ref{fig:fig4}, UniReLo constructs a terrain-relative support
representation, evaluates the resulting support state, and maps its
recovery viability to the continuous progress variable $\alpha_t$.

\paragraph{Terrain-Relative Support Representation.}

Let $\mathcal C_t$ denote the active foot-contact corners. Following
Steps 1--3 in Fig.~\ref{fig:fig4}, their terrain-relative projection,
support representation, and signed support margin are

\begin{equation}
\begin{aligned}
\widetilde{\mathcal C}_{\xi,t}
&=
\Pi_\xi(\mathcal C_t),\\
S_{\xi,t}
&=
\operatorname{conv}
\left(
\widetilde{\mathcal C}_{\xi,t}
\right),\\
m_{\xi,t}^{\mathrm{sup}}
&=
\operatorname{sdist}
\left(
\Pi_\xi(\mathbf p_{\mathrm{com},t}),
S_{\xi,t}
\right).
\end{aligned}
\label{eq:support_region}
\end{equation}

Here, $\Pi_\xi(\cdot)$ projects onto the local terrain tangent plane and
$\operatorname{conv}(\cdot)$ constructs the convex support representation.
The signed margin is positive when the projected CoM lies inside
$S_{\xi,t}$ and negative otherwise.

\paragraph{Support Feasibility Assessment.}

Early recovery may exploit multi-contact support, whereas late recovery
requires a terrain-compatible bipedal configuration. Candidate support
sites include the feet and the representative non-foot sites shown in
Fig.~\ref{fig:fig4}. A site $j$ is considered load bearing according to

\[
c_{j,t}
=
\mathbb I
\left[
(\mathbf f_{j,t}^{w})^\top
\mathbf n_{\xi,t}^{w}
>
f_n^{\min}
\right],
\]

where $\mathbf f_{j,t}^{w}$ is its world-frame contact force and
$\mathbf n_{\xi,t}^{w}$ is the terrain normal. The resulting
contact-redundancy count is
$n_t^c=\sum_{j=1}^{N_s}c_{j,t}$.
The purple markers in Fig.~\ref{fig:fig4} therefore identify candidate
non-foot support sites rather than contacts assumed to remain active
throughout recovery.

Let $h_{\xi,t}$ denote terrain-relative base height and $o_t$ denote
uprightness relative to the terrain normal. The transition from early
multi-contact recovery to late bipedal support is controlled by
$\zeta_t=\operatorname{sigmoid}
(k_h(h_{\xi,t}-h_s)+k_o(o_t-o_s))$.
The knee error $e_{\xi,t}^{\mathrm{knee}}$ is the summed squared distance
of the two knee angles from their terrain-dependent admissible intervals.

Slip and downhill drift are evaluated in the same world-frame terrain
tangent plane. Let
$\mathbf P_{\xi,t}^{\parallel}
=\mathbf I_3-\mathbf n_{\xi,t}^{w}
(\mathbf n_{\xi,t}^{w})^\top$
be the tangent-plane projector, and let
$\mathbf d_{\xi,t}^{w}$ be the normalized projection of gravity onto that
plane. For each foot $b\in\{L,R\}$, $c_{b,t}$ is determined using the same
terrain-normal load criterion as $c_{j,t}$. The motion-stability errors are

\begin{equation}
\begin{aligned}
e_{\mathrm{slip},t}
&=
\frac{
\displaystyle
\sum_{b\in\{L,R\}}
c_{b,t}
\left\|
\mathbf P_{\xi,t}^{\parallel}
\mathbf v_{b,t}^{w}
\right\|_2^2
}{
\displaystyle
\sum_{b\in\{L,R\}}c_{b,t}
+
\varepsilon
},\\
e_{\xi,t}^{\mathrm{down}}
&=
\left[
(\mathbf d_{\xi,t}^{w})^\top
\mathbf P_{\xi,t}^{\parallel}
\mathbf v_{\mathrm{root},t}^{w}
\right]_+ .
\end{aligned}
\label{eq:motion_stability}
\end{equation}

Both terms use velocities expressed in the world frame and projected onto
the same terrain tangent plane. Let $\chi_\xi=1$ for inclined terrain and
$\chi_\xi=0$ otherwise. The support-feasibility reward corresponding to
Step 4 is

\begin{equation}
\begin{aligned}
R_t^{\mathrm{feas}}
={}&
(1-\zeta_t)w_c\log(1+n_t^c)\\
&+
\zeta_t
\left(
w_m m_{\xi,t}^{\mathrm{sup}}
-w_k e_{\xi,t}^{\mathrm{knee}}
-w_s e_{\mathrm{slip},t}
-\chi_\xi w_d e_{\xi,t}^{\mathrm{down}}
\right).
\end{aligned}
\label{eq:feasibility_reward}
\end{equation}

The first term encourages redundant support during early recovery, whereas
the second evaluates locomotion-compatible bipedal support.

\paragraph{Recovery Viability and Progress.}

We collect the support-margin deficit, knee-range violation, foot slip,
terrain-relative height deficit, uprightness deficit, and slope-dependent
downhill drift in a nonnegative error vector $\mathbf e_{\xi,t}$. Its six
entries are
$[m^*-m_{\xi,t}^{\mathrm{sup}}]_+$,
$e_{\xi,t}^{\mathrm{knee}}$,
$e_{\mathrm{slip},t}$,
$[h^*-h_{\xi,t}]_+$,
$[o^*-o_t]_+$, and
$\chi_\xi e_{\xi,t}^{\mathrm{down}}$, respectively. Recovery viability is
represented by the weighted energy

\begin{equation}
E_{\xi,t}
=
\boldsymbol{\beta}^{\top}
\mathbf e_{\xi,t}.
\label{eq:recovery_energy}
\end{equation}

Lower $E_{\xi,t}$ indicates greater compatibility with stable locomotion.
The viable recovery-state set is
$\mathcal R_\xi=
\{\mathbf s:E_\xi(\mathbf s)\leq E_{\mathrm{tol}}\}$.

The guidance reward and recovery progress are defined as

\begin{equation}
\begin{aligned}
R_t^{\mathrm{TCG}}
&=
R_t^{\mathrm{feas}}
+
\lambda_{\mathrm{pot}}
\left(
E_{\xi,t}
-
\gamma E_{\xi,t+1}
\right),\\
\alpha_t
&=
\operatorname{clip}
\left(
1-
\frac{[E_{\xi,t}-E_{\mathrm{tol}}]_+}
{E_{\mathrm{norm}}-E_{\mathrm{tol}}+\varepsilon},
0,1
\right).
\end{aligned}
\label{eq:tcg_progress}
\end{equation}

Thus, $\alpha_t$ increases as the robot approaches a terrain-compatible
state capable of sustaining subsequent locomotion. Recovery progress also
coordinates the transition between rising and command tracking:

\begin{equation}
R_t^{\mathrm{task}}
=
(1-\alpha_t)R_t^{\mathrm{rise}}
+
\alpha_t R_t^{\mathrm{cmd}}.
\label{eq:task_reward}
\end{equation}

Here, $R_t^{\mathrm{rise}}$ rewards terrain-relative base height and
uprightness, whereas $R_t^{\mathrm{cmd}}$ rewards tangent-plane
linear-velocity and yaw-rate tracking. The same $\alpha_t$ controls the
multi-scale motion-prior gates in Eq.~\eqref{eq:gates}.

\begin{table}[!t]
\centering
\scriptsize
\setlength{\tabcolsep}{6pt}
\renewcommand{\arraystretch}{1.04}
\caption{Task and regularization weights used in UniReLo.}
\label{tab:reward_summary}
\begin{tabular}{lc}
\toprule
\textbf{Term} & \textbf{Weight}\\
\midrule
Terrain-relative height             & $4.5$\\
Body uprightness                    & $1.0$\\
Linear-velocity tracking            & $0.65$\\
Yaw-rate tracking                   & $1.0$\\
Action variation                    & $0.015$\\
Joint velocity                      & $2.5{\times}10^{-7}$\\
Joint torque                        & $2.5{\times}10^{-6}$\\
Joint-limit violation               & $100$\\
Undesired collision                 & $1.0$\\
\bottomrule
\end{tabular}
\end{table}

\subsection{Terrain-Pose Plasticity-Aware Initialization}
\label{subsec:tpp}

Uniform initialization may overrepresent trivial poses or infeasible
terrain--pose pairs. UniReLo therefore prioritizes fallen configurations
whose recoverability is sensitive to terrain conditions.

A fallen-pose template is represented as
$\boldsymbol{\psi}_j=
[\mathbf q_j,\mathbf g_j^b,\theta_j^{\mathrm{rel}},
\mathbf v_{0,j},\boldsymbol{\omega}_{0,j}]$,
where $\theta_j^{\mathrm{rel}}$ denotes the body orientation relative to the
local terrain surface. Its terrain-dependent plasticity is measured by

\begin{equation}
I_j
=
\operatorname{Var}_{\xi\in\mathcal K}
\left(
\hat p_{j,\xi}
\right),
\label{eq:plasticity_score}
\end{equation}

where $\hat p_{j,\xi}$ is the empirical recovery-success probability.
A larger $I_j$ indicates stronger terrain-dependent variation in
recoverability.

Terrain--pose pairs are sampled according to

\begin{equation}
P(j,\xi)
\propto
\lambda_v(1-\hat V_{j,\xi})
+
\lambda_i I_j
+
\lambda_u(N_{j,\xi}+1)^{-1/2},
\label{eq:sampling_priority}
\end{equation}

where $\hat V_{j,\xi}\in[0,1]$ is the normalized moving-average return and $N_{j,\xi}$ is the visitation count. This distribution emphasizes
challenging, terrain-sensitive, and underexplored initial states.

All terrain-dependent quantities and motion discriminators are used only during training; deployment retains the proprioceptive actor and low-level joint-position controller.

\begin{figure}[t]
\centering
\begin{minipage}[t]{0.48\linewidth}
  \centering
  \includegraphics[width=\linewidth]{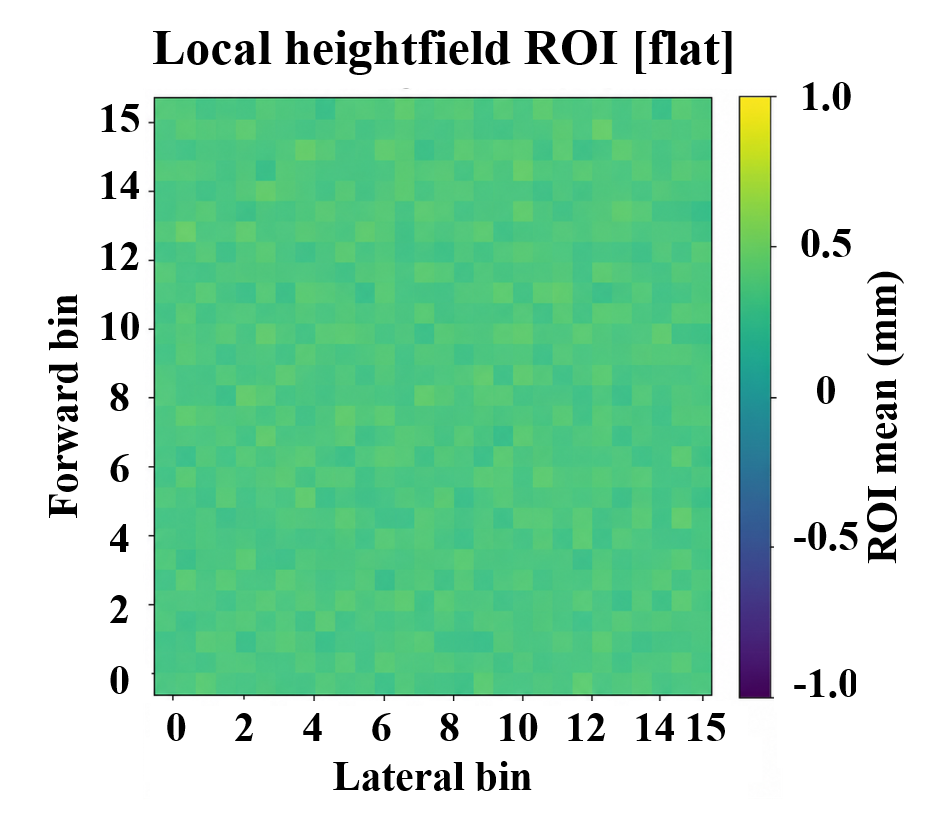}
  \subcaption{Flat ground.}
\end{minipage}%
\hfill
\begin{minipage}[t]{0.49\linewidth}
  \centering
  \includegraphics[width=\linewidth]{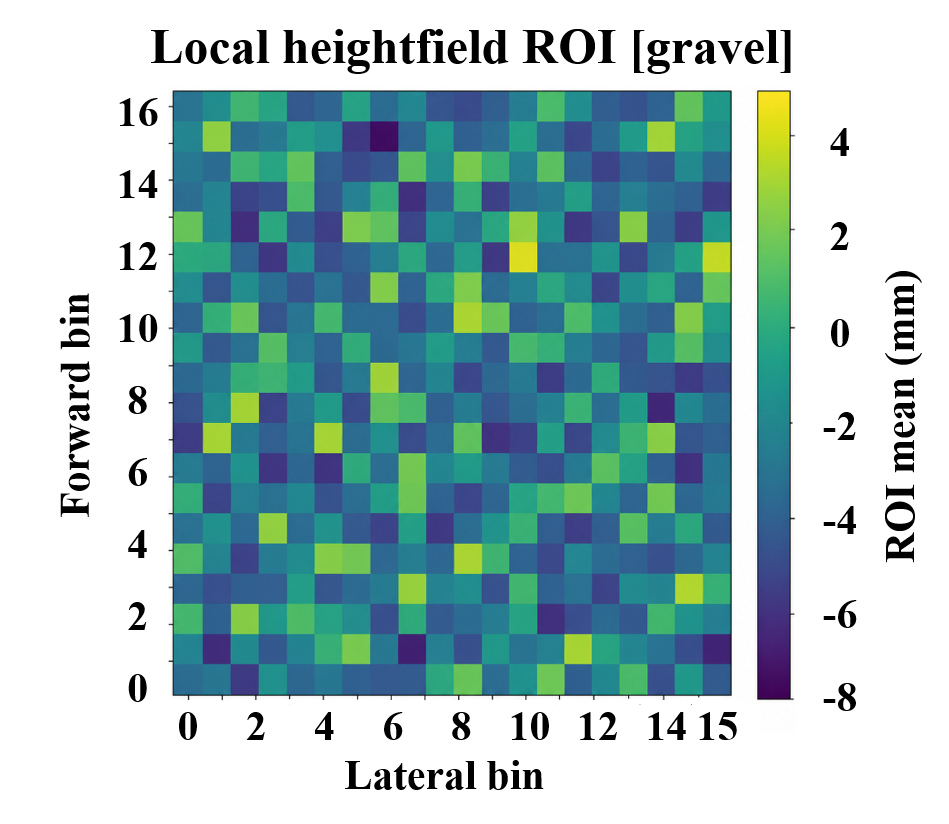}
  \subcaption{Gravel.}
\end{minipage}

\vspace{0.5em}

\begin{minipage}[t]{0.49\linewidth}
  \centering
  \includegraphics[width=\linewidth]{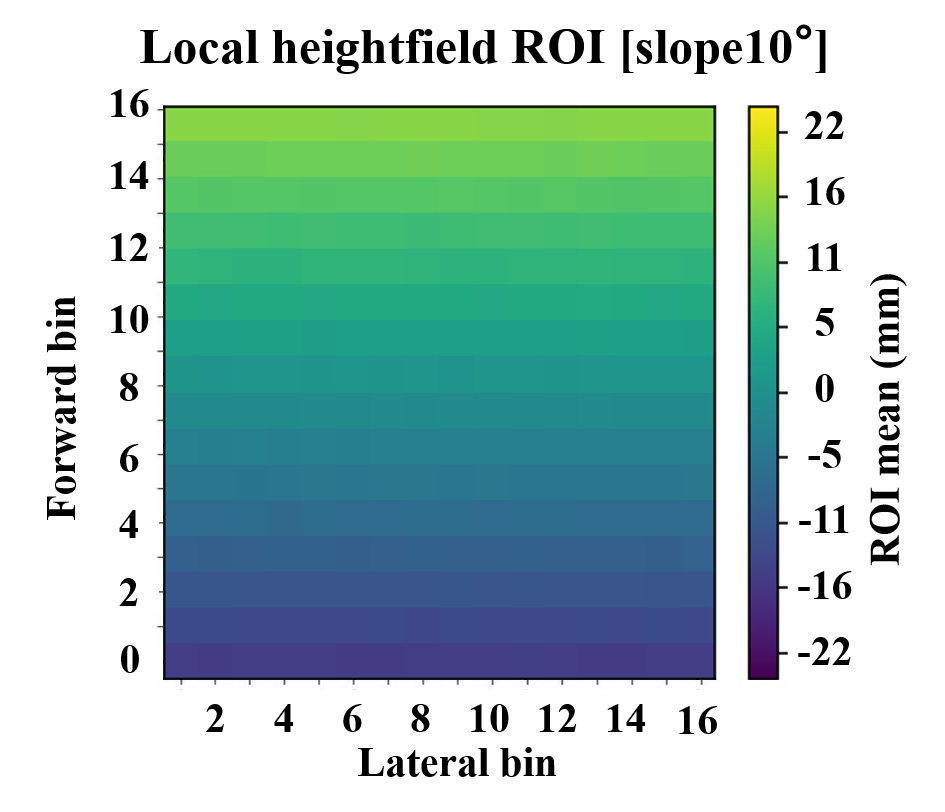}
  \subcaption{Slope $10^\circ$.}
\end{minipage}%
\hfill
\begin{minipage}[t]{0.49\linewidth}
  \centering
  \includegraphics[width=\linewidth]{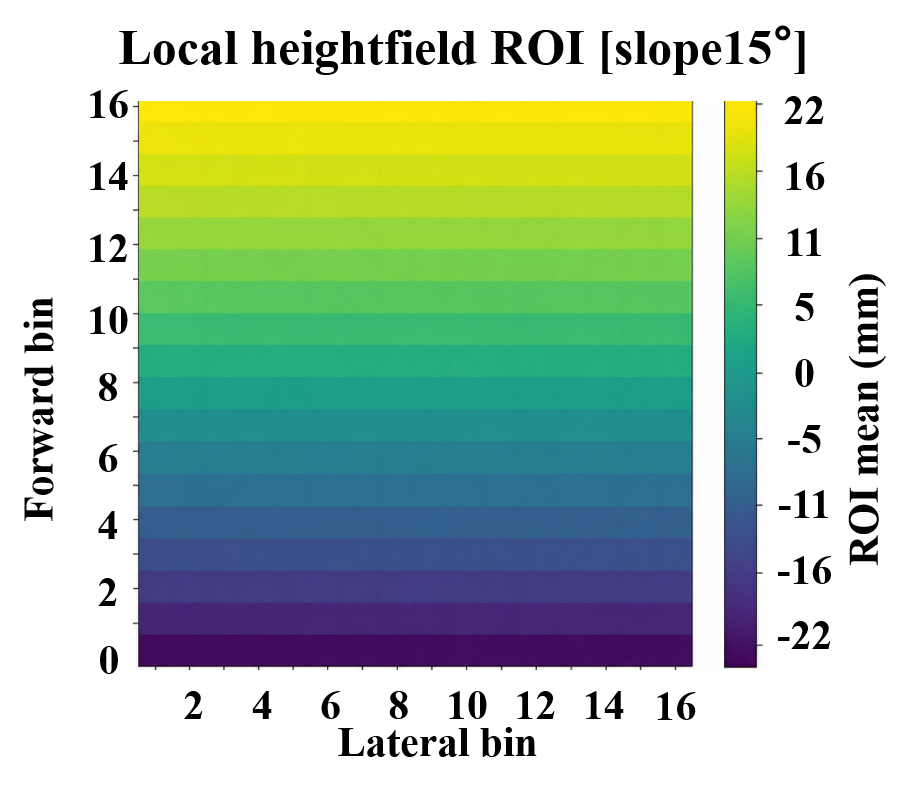}
  \subcaption{Slope $15^\circ$.}
\end{minipage}

\caption{
Elevation distributions within the foot-contact regions of the four
simulated terrains. Elevations are expressed relative to the mean of each
local region. Flat ground exhibits nearly uniform elevation, gravel contains
stochastic local variations, and the two inclined terrains exhibit
consistent forward elevation gradients. The $10^\circ$ and $15^\circ$
slope panels use the same color range to enable direct comparison of their
gradient magnitudes.
}
\label{fig:terrain_heatmaps}
\end{figure}

    \section{EXPERIMENTS}
    \label{sec:experiments}
    
    We evaluate UniReLo from four perspectives: recovery robustness across
    representative field terrains, terrain-conditioned recovery behavior,
    the contribution of the proposed components, and sim-to-real transfer to
    outdoor environments. For fair comparison with existing standing-up
    methods, the baseline evaluation adopts a zero-command get-up-to-stand
    protocol. The ablation and hardware evaluations further use continuous
    commands to assess the complete transition from fallen states to sustained
    locomotion.
    
    \subsection{Experimental Settings}
    \label{subsec:exp_settings}
    
    \paragraph{Simulation and Terrain Setup.}
    
\begin{figure}[t]
    \centering
    \includegraphics[width=\linewidth]{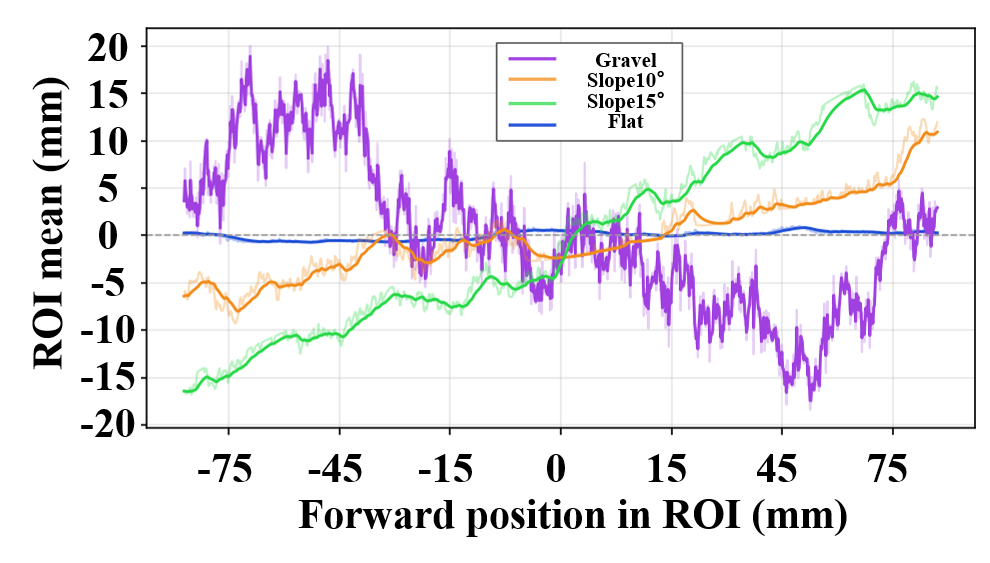}
    \caption{
    Longitudinal elevation profiles obtained by averaging the mean-centered
    local heightfield along the lateral direction. The horizontal axis denotes
    the forward position within the foot-contact region. The vertical axis,
    labeled \emph{ROI mean}, denotes the lateral mean elevation at each forward
    position relative to the mean elevation of the complete ROI. Flat ground
    remains near zero, gravel exhibits irregular local fluctuations, and the
    inclined terrains produce approximately linear profiles whose gradients
    increase with slope angle.
    }
    \label{fig:roi_profile}
    \end{figure}

    Training and evaluation are conducted on flat ground, gravel, and planar
    slopes of $10^\circ$ and $15^\circ$. These terrains represent rigid
    structured support, locally irregular contact, and increasingly inclined
    support conditions, respectively. Their elevation distributions and
    longitudinal profiles within the local foot-contact region are shown in
    Figs.~\ref{fig:terrain_heatmaps} and~\ref{fig:roi_profile}.
    
    \paragraph{Implementation Details.}
    
    UniReLo and its ablated variants are trained using
    PPO~\cite{schulman2017ppo} with 4,096 parallel environments at a control
    frequency of 50\,Hz. The policy input contains a five-frame proprioceptive
    history. The PPO optimizer uses a learning rate of $3\times10^{-4}$, a
    clipping ratio of $0.2$, a GAE parameter of $\lambda=0.95$, and a discount
    factor of $\gamma=0.99$.
    
    Domain randomization is applied to link masses, joint damping, contact
    friction, observation noise, communication delay, external disturbances,
    and terrain properties. On gravel, surface-height perturbations are sampled
    within $\pm0.02$\,m and friction coefficients are sampled from
    $\mathcal{U}(0.3,0.6)$. The inclined terrains are planar surfaces with fixed
    inclination angles.
    
    \paragraph{Baseline Evaluation Protocol and Metrics.}
    \label{subsec:comparison}
    
    For comparison with existing humanoid fall-recovery methods, all policies
    are evaluated under a unified get-up-to-stand protocol with zero linear and
    angular velocity commands.
    This protocol evaluates whether the robot can recover and maintain a stable
    upright state without requiring command-following locomotion. We report
    Success Rate (SR), Time-to-Stand (TTS), and Time-to-Fall (TTF).
    
    \begin{enumerate}
    
    \item \textbf{Success Rate (SR, \%):}
    The percentage of trials in which the robot recovers from the fallen state
    and reaches the prescribed stable-upright condition.
    
    \item \textbf{Time-to-Stand (TTS, s):}
    The elapsed time from policy activation until the stable-upright condition
    is first established. TTS is evaluated only over successful trials.
    
    \item \textbf{Time-to-Fall (TTF, s):}
    The elapsed time from the first stable-upright instant to a secondary fall.
    TTF is evaluated only for trials in which the robot first reaches the
    upright condition and subsequently loses balance.
    
    \end{enumerate}
    
    We compare UniReLo with HumanUP~\cite{he2025getup},
    HoST~\cite{huang2025host}, and FIRM~\cite{xu2025firm}. All methods are
    implemented and evaluated using the same full 29-DoF Unitree G1 rigid-body
    model. To preserve their original control formulations, each baseline
    retains the actively controlled joint set adopted in its original
    implementation. Joints outside a baseline's policy interface are held at
    their nominal positions through low-level PD control.
    
    The baseline-specific network architectures, learning objectives, reward
    organizations, and training procedures are retained according to their
    original formulations. Only simulator-dependent interfaces and
    robot-specific mappings are adapted to the common platform. During
    evaluation, all methods use the same terrain configurations, balanced
    initial-posture distributions, episode horizon, stable-upright criterion,
    and evaluation procedure.
    
    All methods are evaluated over prone, supine, and side initial postures.
    Results are reported as the mean and standard deviation over five
    independent random seeds, with 200 trials conducted for each
    posture--terrain combination under every seed.
    
    \subsection{Simulation Evaluation}
    \label{subsec:sim_analysis}
    
    \paragraph{Comparison with Existing Recovery Methods.}
    
    Table~\ref{tab:comparison_sim} reports aggregate results under the unified
    zero-command get-up-to-stand protocol. On flat ground, all methods achieve
    relatively strong recovery performance. HoST obtains the shortest TTS,
    whereas FIRM and UniReLo exhibit no secondary fall within the evaluation
    horizon.
    
    The differences become more pronounced as terrain complexity increases.
    FIRM performs best among the prior methods on gravel, while HoST remains the
    strongest baseline on inclined terrain. UniReLo nevertheless achieves the
    highest SR under every evaluated terrain condition. Relative to the strongest
    baseline on each terrain, UniReLo provides relative SR improvements of
    $8.3\%$ on gravel, $11.8\%$ on the $10^\circ$ slope, and $69.0\%$ on the
    $15^\circ$ slope. Its TTS remains within $1$\,s of the fastest method on
    flat ground and is the lowest on gravel and both slopes.
    
    On the $15^\circ$ slope, representative unsuccessful baseline trials reach
    an approximately upright posture but subsequently exhibit foot sliding or
    downhill drift. In contrast, UniReLo maintains substantially higher recovery
    success and longer post-standing stability. These results support evaluating
    the terrain compatibility of the recovered support state rather
    than body elevation alone.
    
    \begin{table*}[!t]
    \centering
    \tiny
    \setlength{\tabcolsep}{2pt}
    \renewcommand{\arraystretch}{1.15}
    \caption{
    Simulation comparison with existing humanoid recovery methods under the
    zero-command get-up-to-stand protocol.
    }
    \label{tab:comparison_sim}
    
    \resizebox{\linewidth}{!}{
    \begin{tabular}{lcccccccccccc}
    \toprule
    \multirow{2}{*}{Method}
    & \multicolumn{3}{c}{Flat}
    & \multicolumn{3}{c}{Gravel}
    & \multicolumn{3}{c}{Slope $10^\circ$}
    & \multicolumn{3}{c}{Slope $15^\circ$} \\
    \cmidrule(lr){2-4}
    \cmidrule(lr){5-7}
    \cmidrule(lr){8-10}
    \cmidrule(lr){11-13}
    & SR$\uparrow$
    & TTS$\downarrow$
    & TTF$\uparrow$
    & SR$\uparrow$
    & TTS$\downarrow$
    & TTF$\uparrow$
    & SR$\uparrow$
    & TTS$\downarrow$
    & TTF$\uparrow$
    & SR$\uparrow$
    & TTS$\downarrow$
    & TTF$\uparrow$ \\
    \midrule
    
    HoST~\cite{huang2025host}
    & $92.12_{\pm0.68}$
    & $\mathbf{1.82_{\pm0.06}}$
    & $0.15_{\pm0.21}$
    & $23.45_{\pm1.62}$
    & $3.28_{\pm0.74}$
    & $1.41_{\pm1.18}$
    & $83.40_{\pm2.35}$
    & $2.98_{\pm0.42}$
    & $1.85_{\pm0.55}$
    & $52.91_{\pm2.47}$
    & $3.68_{\pm0.61}$
    & $0.92_{\pm0.48}$ \\
    
    FIRM~\cite{xu2025firm}
    & $90.56_{\pm0.82}$
    & $2.58_{\pm0.54}$
    & N/A
    & $89.74_{\pm1.35}$
    & $3.05_{\pm0.91}$
    & $1.62_{\pm1.06}$
    & $57.20_{\pm2.48}$
    & $3.55_{\pm0.72}$
    & $1.12_{\pm0.86}$
    & $40.30_{\pm2.98}$
    & $4.28_{\pm0.83}$
    & $0.68_{\pm0.59}$ \\
    
    HumanUP~\cite{he2025getup}
    & $89.85_{\pm0.48}$
    & $6.18_{\pm0.26}$
    & $0.37_{\pm0.19}$
    & $61.30_{\pm1.78}$
    & $6.84_{\pm0.65}$
    & $0.88_{\pm0.72}$
    & $53.83_{\pm2.27}$
    & $6.72_{\pm0.48}$
    & $0.78_{\pm0.42}$
    & $15.20_{\pm2.61}$
    & $6.35_{\pm0.72}$
    & $0.35_{\pm0.28}$ \\
    
    \textbf{UniReLo (Ours)}
    & $\mathbf{99.53_{\pm0.35}}$
    & $2.35_{\pm0.08}$
    & N/A
    & $\mathbf{97.18_{\pm0.33}}$
    & $\mathbf{2.13_{\pm0.93}}$
    & N/A
    & $\mathbf{93.28_{\pm0.31}}$
    & $\mathbf{2.76_{\pm0.32}}$
    & $\mathbf{3.06_{\pm0.40}}$
    & $\mathbf{89.42_{\pm0.26}}$
    & $\mathbf{3.57_{\pm0.79}}$
    & $\mathbf{2.14_{\pm0.63}}$ \\
    
    \bottomrule
    \end{tabular}
    }
    
    \vspace{0.4em}
    \scriptsize
    ``N/A'' indicates that no secondary fall occurred within the evaluation
    horizon.
    \end{table*}
    
    \paragraph{Recovery Progress and Terrain-Dependent Signals.}
    
    \begin{figure}[t]
    \centering
    \includegraphics[width=\linewidth]{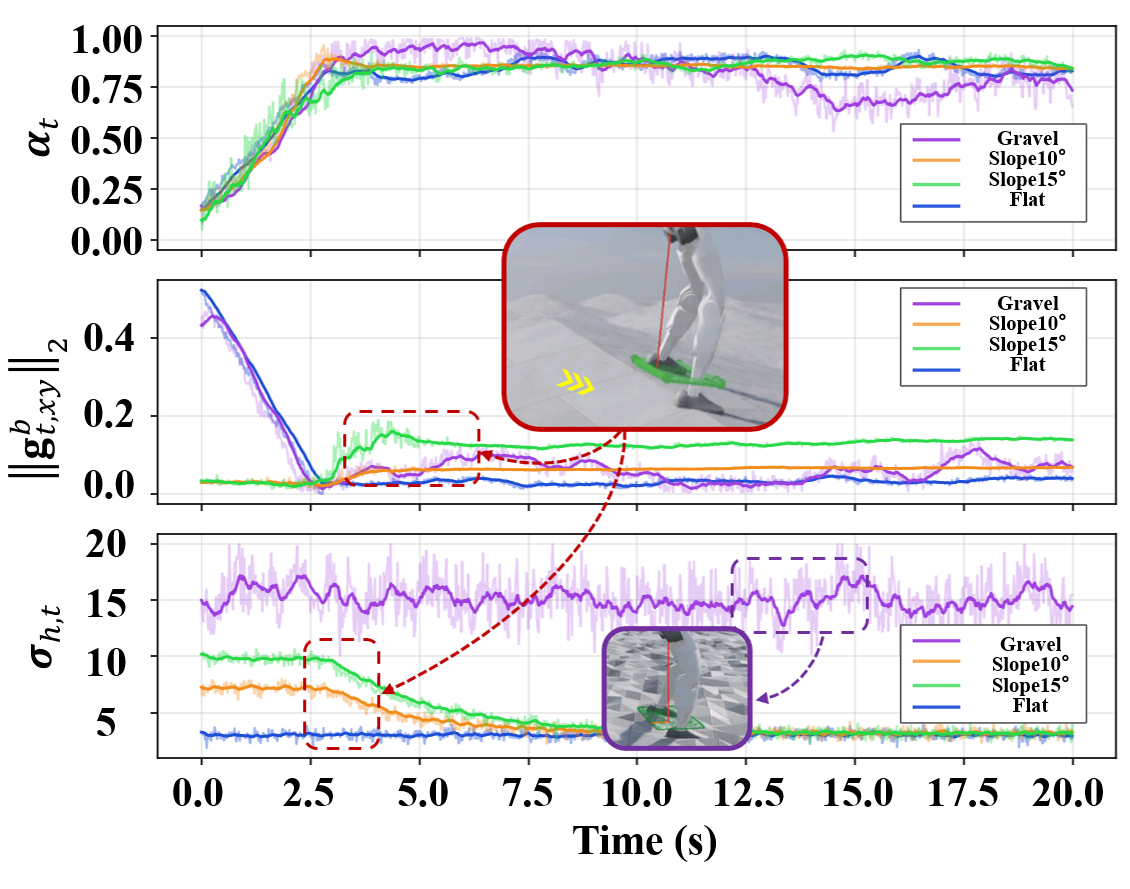}
    \caption{
    Temporal evolution of recovery progress $\alpha_t$, base-tilt magnitude
    $\|\mathbf g^{b}_{t,xy}\|_2$, and local contact-region height variation
    $\sigma_{h,t}$ across flat ground, gravel, and the $10^\circ$ and
    $15^\circ$ slopes. Solid curves denote mean trajectories, shaded regions
    indicate variability, and the insets show representative contact
    configurations. The terrain-height statistic is used only for analysis and
    is not part of the deployed actor observation.
    }
    \label{fig:roi_triple}
    \end{figure}
    
    Fig.~\ref{fig:roi_triple} examines the recovery dynamics under the four
    terrain conditions. The recovery progress $\alpha_t$, produced by the
    terrain-conditioned recovery guidance, increases as the robot approaches a
    terrain-compatible locomotion-ready state. Meanwhile, the base-tilt
    magnitude decreases as the whole-body posture and contact configuration are
    reorganized.
    
    The inclined terrains retain nonzero steady-state tilt because the recovered
    posture aligns with the terrain rather than with a nominal flat-ground
    configuration. The local height statistic $\sigma_{h,t}$ distinguishes the
    underlying terrain structures: flat ground remains nearly uniform, gravel
    contains stochastic local variation, and planar slopes produce variation
    that increases with inclination. Here, $\sigma_{h,t}$ is a diagnostic
    terrain statistic rather than the direct definition of $\alpha_t$; recovery
    progress is determined by the combined support, posture, and motion
    feasibility introduced in Sec.~\ref{subsec:tcg}.
    
    \paragraph{Terrain-Compatible Recovery Behaviors.}
    
    \begin{figure*}[!t]
    \centering
    
    \begin{subfigure}{\textwidth}
    \centering
    \includegraphics[width=\linewidth]{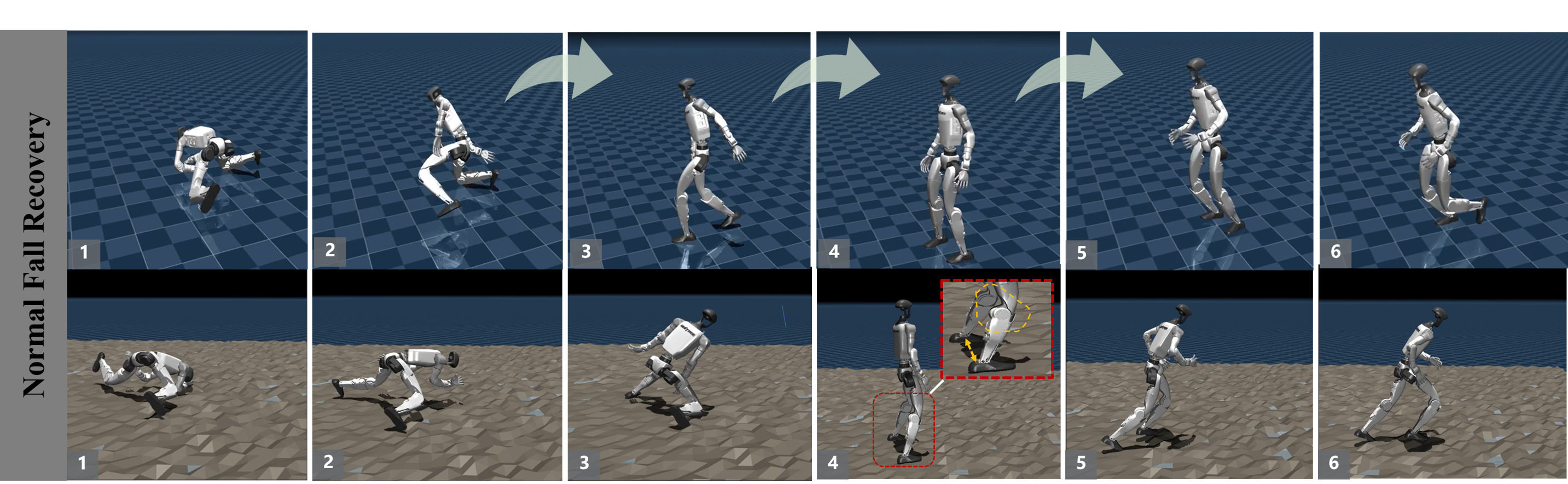}
    \caption{Flat ground (top) and gravel (bottom).}
    \label{fig:success_a}
    \end{subfigure}
    
    \begin{subfigure}{\textwidth}
    \centering
    \includegraphics[width=\linewidth]{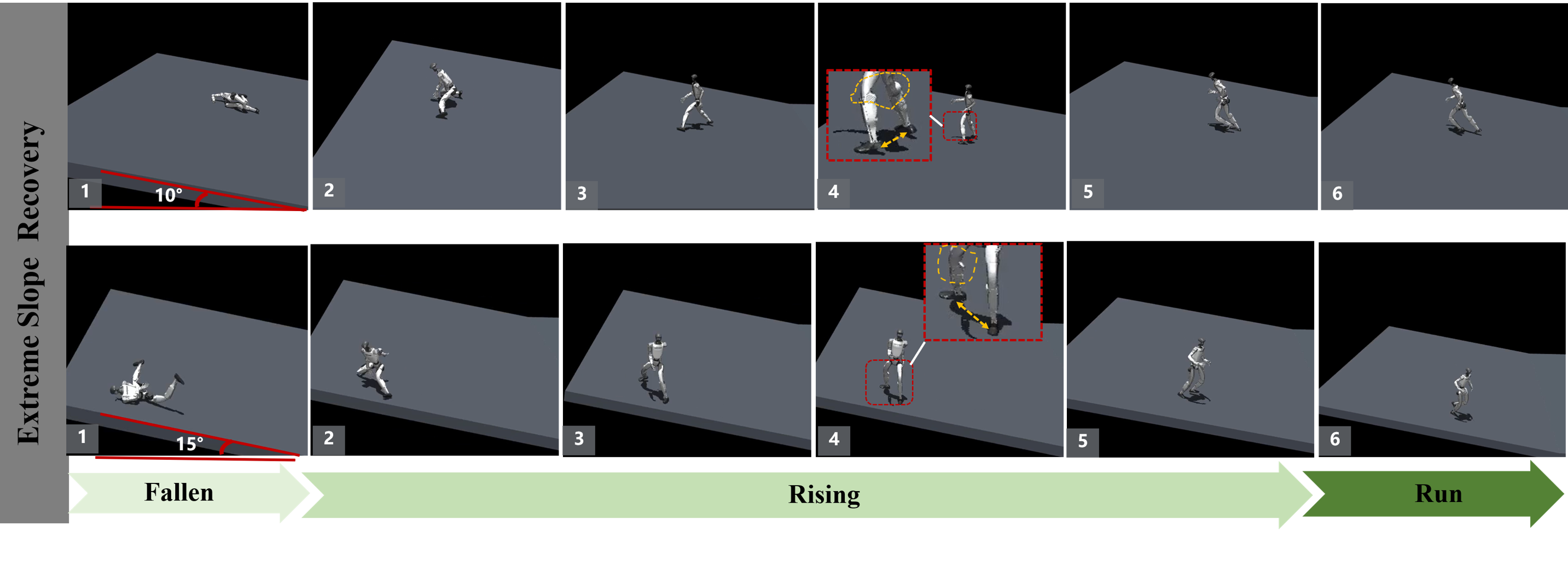}
    \caption{$10^\circ$ slope (top) and $15^\circ$ slope (bottom).}
    \label{fig:success_b}
    \end{subfigure}
    
    \caption{
    Qualitative recovery and command-following sequences across heterogeneous
    simulated terrains. On gravel, the red boxes highlight persistent
    foot-placement adjustments under irregular contact. On inclined terrain,
    the highlighted configurations show increasing knee flexion and
    terrain-aligned whole-body adaptation as the slope angle increases. The
    background stage labels are included only for visualization and do not
    represent internal policy modes or explicit switching conditions.
    }
    \label{fig:success_sequences}
    \end{figure*}
    \begin{figure*}[!t]
    \centering
    
    \begin{subfigure}{0.49\textwidth}
    \centering
    \includegraphics[width=\linewidth]{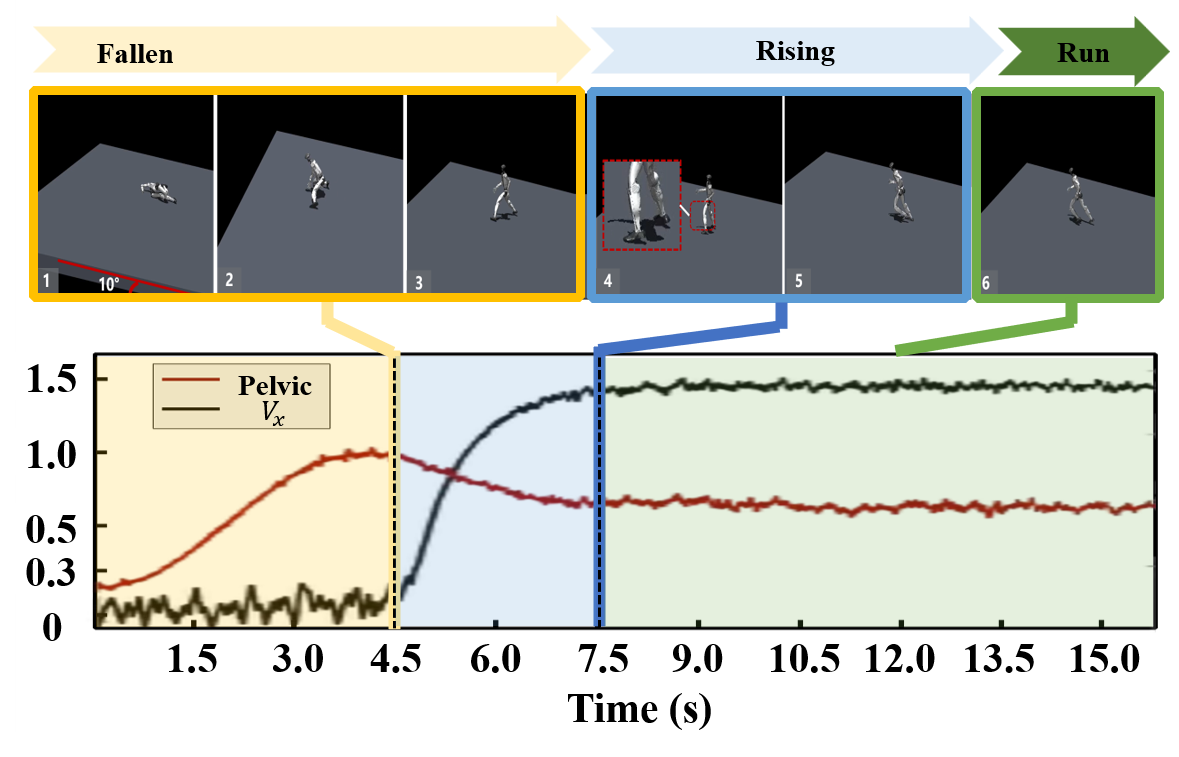}
    \caption{UniReLo.}
    \label{fig:tprcs_poses_a}
    \end{subfigure}
    \hfill
    \begin{subfigure}{0.48\textwidth}
    \centering
    \includegraphics[width=\linewidth]{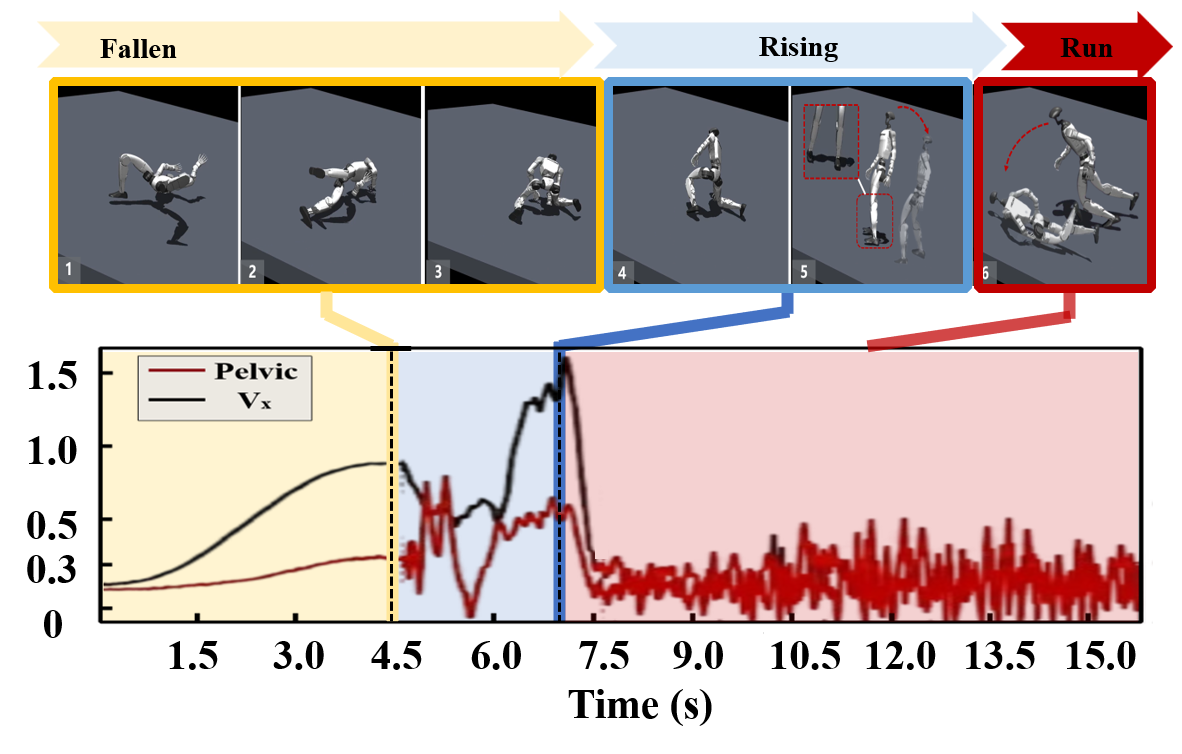}
    \caption{Hard-gated Routing.}
    \label{fig:tprcs_poses_b}
    \end{subfigure}
    
    \caption{
    Continuous gating versus Hard-gated Routing on a $10^\circ$ slope under
    the continuous-command protocol. UniReLo maintains smooth forward-velocity
    and pelvic-pitch evolution from recovery to sustained command tracking.
    Hard-gated Routing introduces an abrupt dominant-prior change, followed by
    velocity and pelvic-pitch oscillations and eventual instability. The marked
    region indicates the routing change of the hard-gated variant. The colored
    backgrounds denote visualization-only behavior intervals and do not
    represent internal modes of UniReLo.
    }
    \label{fig:gate_comparison}
    \end{figure*}
    Fig.~\ref{fig:success_sequences} shows that the recovered configuration
    changes with terrain instead of converging to one nominal standing posture.
    Flat ground permits a relatively direct transition into locomotion. On
    gravel, the policy performs persistent foot-placement adjustments and
    maintains a more compliant posture to accommodate irregular support. On
    slopes, the robot increases knee flexion and aligns its body configuration
    with the terrain gradient, with stronger adaptation on the $15^\circ$
    slope.
    
    These behaviors are consistent with the terrain-dependent knee configuration and
    support-feasibility assessment used by the terrain-conditioned recovery
    guidance. Recovery therefore targets a terrain-compatible support
    configuration from which command-following locomotion can be sustained.
    
    \subsection{Ablation Study}
    \label{subsec:ablation}
    
    \paragraph{Variants and Evaluation Protocol.}
    
    All ablation variants are evaluated with a constant forward-velocity command
    of $0.25$\,m/s and zero lateral-velocity and yaw-rate commands. Unlike the
    zero-command baseline protocol, this setting evaluates the complete process
    from fall recovery to sustained command tracking.
    
    We evaluate four variants:
    
    \begin{enumerate}
    
    \item \textbf{w/o TCG (without terrain-conditioned recovery guidance)}, which removes
    the reward-shaping branch of terrain-conditioned recovery guidance,
    including $R_t^{\mathrm{feas}}$ and the potential-based energy-reduction
    term. The recovery energy $E_{\xi,t}$ is retained only for computing
    $\alpha_t$, thereby preserving task blending and continuous motion-prior
    gating.
    
    \item \textbf{w/o TPP Init.}, which replaces terrain-pose
    plasticity-aware initialization with uniform sampling over terrains and
    fallen-pose templates.
    
    \item \textbf{Single-scale AMP}, which replaces the multi-scale
    discriminator bank with one AMP discriminator trained on the complete
    recovery-to-locomotion sequence without temporal-scale decomposition.
    
    \item \textbf{Hard-gated Routing}, which replaces continuous
    progress-dependent blending with one-hot prior selection:
    \begin{equation}
    \label{eq:hard_gate}
    \widetilde g_i(\alpha_t)
    =
    \mathbb{I}
    \left[
    i=\arg\max_j g_j(\alpha_t)
    \right],
    \end{equation}
the one-hot weights replace $g_i(\alpha_t)$ in both motion-prior reward
fusion and policy-sample weighting during discriminator training.
    
    \end{enumerate}
    
    The baseline comparison in Table~\ref{tab:comparison_sim} evaluates
    get-up-to-stand performance under a zero command. In contrast, the ablation
    study adopts a stricter recovery-to-tracking criterion under a continuous
    command.
    
    We report the \emph{Recovery-to-Tracking Success Rate} (RTSR). A trial is
    successful only if the policy recovers, establishes sustained command
    tracking, and avoids a secondary fall within the evaluation horizon. We also
    report the \emph{Average Signed Support Margin} (ASM) and the
    \emph{Command Tracking Error} (CTE):
    \begin{equation}
    \label{eq:ablation_metrics}
    \begin{aligned}
    \mathrm{ASM}
    &=
    \frac{1}{|\mathcal T_u|}
    \sum_{t\in\mathcal T_u}
    m_{\xi,t}^{\mathrm{sup}},
    \\
    \mathrm{CTE}
    &=
    \frac{1}{|\mathcal T_c|}
    \sum_{t\in\mathcal T_c}
    \left\|
    \begin{bmatrix}
    (v_{x,t}-v_{x,t}^{\mathrm{cmd}})/v_{\mathrm{ref}}
    \\
    (v_{y,t}-v_{y,t}^{\mathrm{cmd}})/v_{\mathrm{ref}}
    \\
    (\omega_{z,t}-\omega_{z,t}^{\mathrm{cmd}})/\omega_{\mathrm{ref}}
    \end{bmatrix}
    \right\|_2 .
    \end{aligned}
    \end{equation}
    Here, $\mathcal T_u$ denotes the interval after the first stable-upright
    instant, whereas $\mathcal T_c$ denotes the subsequent command-evaluation
    interval. ASM measures terrain-relative support robustness and is reported
    in centimeters. CTE measures normalized planar-velocity and yaw-rate
    tracking errors. ASM and CTE are computed only for trials reaching the
    corresponding evaluation intervals, while unsuccessful trials are reflected
    by RTSR.

    \paragraph{Effect of Continuous Multi-Scale Gating.}
    
    Having examined how terrain-conditioned recovery guidance produces
    $\alpha_t$, we next evaluate how this shared progress variable coordinates
    the multi-scale motion priors. Fig.~\ref{fig:gate_comparison} compares
    UniReLo and Hard-gated Routing from the same prone configuration on a
    $10^\circ$ slope.
    
    UniReLo smoothly changes the contributions of the frame-, sequence-, and
    gait-level priors as $\alpha_t$ evolves. In contrast, discrete prior
    replacement creates a discontinuity near the routing boundary, followed by
    pronounced forward-velocity and pelvic-pitch oscillations and a secondary
    fall. This result directly illustrates the transition instability that
    continuous progress-dependent gating is designed to avoid.
    \begin{table*}[t]
    \centering
    \tiny
    \setlength{\tabcolsep}{3pt}
    \renewcommand{\arraystretch}{1.1}
    
    \caption{
    Posture-resolved component ablation across four terrains under the
    continuous-command evaluation protocol. Values are reported as the mean
    and standard deviation over five random seeds, with 200 trials conducted
    for each posture--terrain combination under every seed.
    }
    \label{tab:ablation_all}
    
    \resizebox{\linewidth}{!}{
    \begin{tabular}{lcccccccccccc}
    \toprule
    
    \multirow{2}{*}{Variant}
    & \multicolumn{3}{c}{Flat}
    & \multicolumn{3}{c}{Gravel}
    & \multicolumn{3}{c}{Slope $10^\circ$}
    & \multicolumn{3}{c}{Slope $15^\circ$} \\
    
    \cmidrule(lr){2-4}
    \cmidrule(lr){5-7}
    \cmidrule(lr){8-10}
    \cmidrule(lr){11-13}
    
    & RTSR$\uparrow$
    & ASM$\uparrow$
    & CTE$\downarrow$
    & RTSR$\uparrow$
    & ASM$\uparrow$
    & CTE$\downarrow$
    & RTSR$\uparrow$
    & ASM$\uparrow$
    & CTE$\downarrow$
    & RTSR$\uparrow$
    & ASM$\uparrow$
    & CTE$\downarrow$ \\
    
    \midrule
    
    \multicolumn{13}{l}{\textbf{(1) Prone Initial Postures}} \\
    \cmidrule{1-13}
    
    w/o TCG
    & $96.4_{\pm 0.89}$
    & $2.94_{\pm 0.31}$
    & $0.22_{\pm 0.08}$
    & $89.8_{\pm 1.86}$
    & $2.11_{\pm 0.11}$
    & $0.33_{\pm 0.04}$
    & $81.3_{\pm 1.57}$
    & $1.58_{\pm 0.19}$
    & $0.39_{\pm 0.12}$
    & $71.5_{\pm 0.91}$
    & $0.87_{\pm 0.28}$
    & $0.44_{\pm 0.08}$ \\
    \cmidrule{1-13}
    
    w/o TPP Init
    & $93.2_{\pm 1.72}$
    & $3.15_{\pm 0.25}$
    & $0.19_{\pm 0.03}$
    & $84.4_{\pm 0.65}$
    & $2.41_{\pm 0.16}$
    & $0.29_{\pm 0.08}$
    & $77.2_{\pm 1.58}$
    & $2.11_{\pm 0.38}$
    & $0.34_{\pm 0.11}$
    & $66.3_{\pm 1.57}$
    & $1.54_{\pm 0.25}$
    & $0.39_{\pm 0.10}$ \\
    \cmidrule{1-13}
    
    Single-scale AMP
    & $96.1_{\pm 1.50}$
    & $3.19_{\pm 0.32}$
    & $0.18_{\pm 0.02}$
    & $92.3_{\pm 0.71}$
    & $2.58_{\pm 0.09}$
    & $0.27_{\pm 0.06}$
    & $84.1_{\pm 1.34}$
    & $2.30_{\pm 0.16}$
    & $0.32_{\pm 0.06}$
    & $75.2_{\pm 2.00}$
    & $1.71_{\pm 0.24}$
    & $0.36_{\pm 0.02}$ \\
    \cmidrule{1-13}
    
    Hard-gated Routing
    & $97.9_{\pm 0.76}$
    & $3.26_{\pm 0.42}$
    & $0.17_{\pm 0.06}$
    & $93.6_{\pm 1.15}$
    & $2.63_{\pm 0.29}$
    & $0.26_{\pm 0.02}$
    & $85.7_{\pm 1.88}$
    & $2.32_{\pm 0.34}$
    & $0.30_{\pm 0.07}$
    & $77.0_{\pm 1.17}$
    & $1.77_{\pm 0.42}$
    & $0.35_{\pm 0.05}$ \\
    \cmidrule{1-13}
    
    \textbf{Full UniReLo}
    & $\mathbf{99.6_{\pm 0.22}}$
    & $\mathbf{3.50_{\pm 0.28}}$
    & $\mathbf{0.14_{\pm 0.04}}$
    & $\mathbf{98.4_{\pm 0.35}}$
    & $\mathbf{2.97_{\pm 0.20}}$
    & $\mathbf{0.18_{\pm 0.03}}$
    & $\mathbf{94.8_{\pm 0.45}}$
    & $\mathbf{2.64_{\pm 0.24}}$
    & $\mathbf{0.23_{\pm 0.02}}$
    & $\mathbf{89.1_{\pm 0.45}}$
    & $\mathbf{2.08_{\pm 0.18}}$
    & $\mathbf{0.27_{\pm 0.09}}$ \\
    
    \midrule
    
    \multicolumn{13}{l}{\textbf{(2) Supine Initial Postures}} \\
    \cmidrule{1-13}
    
    w/o TCG
    & $95.8_{\pm 0.57}$
    & $2.75_{\pm 0.16}$
    & $0.23_{\pm 0.03}$
    & $87.9_{\pm 1.43}$
    & $1.96_{\pm 0.17}$
    & $0.36_{\pm 0.08}$
    & $79.5_{\pm 1.84}$
    & $1.38_{\pm 0.10}$
    & $0.44_{\pm 0.08}$
    & $70.2_{\pm 1.14}$
    & $0.58_{\pm 0.11}$
    & $0.49_{\pm 0.12}$ \\
    \cmidrule{1-13}
    
    w/o TPP Init
    & $92.6_{\pm 0.96}$
    & $2.91_{\pm 0.35}$
    & $0.21_{\pm 0.05}$
    & $82.6_{\pm 0.55}$
    & $2.25_{\pm 0.39}$
    & $0.32_{\pm 0.12}$
    & $74.8_{\pm 1.40}$
    & $1.94_{\pm 0.29}$
    & $0.38_{\pm 0.11}$
    & $65.1_{\pm 1.39}$
    & $1.38_{\pm 0.37}$
    & $0.44_{\pm 0.09}$ \\
    \cmidrule{1-13}
    
    Single-scale AMP
    & $95.4_{\pm 1.52}$
    & $2.98_{\pm 0.12}$
    & $0.20_{\pm 0.07}$
    & $90.3_{\pm 0.67}$
    & $2.41_{\pm 0.15}$
    & $0.30_{\pm 0.11}$
    & $82.9_{\pm 1.75}$
    & $2.10_{\pm 0.09}$
    & $0.37_{\pm 0.09}$
    & $73.6_{\pm 0.84}$
    & $1.53_{\pm 0.34}$
    & $0.41_{\pm 0.07}$ \\
    \cmidrule{1-13}
    
    Hard-gated Routing
    & $96.2_{\pm 1.47}$
    & $3.05_{\pm 0.29}$
    & $0.19_{\pm 0.04}$
    & $91.4_{\pm 0.89}$
    & $2.46_{\pm 0.11}$
    & $0.28_{\pm 0.06}$
    & $83.4_{\pm 1.35}$
    & $2.17_{\pm 0.21}$
    & $0.31_{\pm 0.05}$
    & $75.8_{\pm 1.04}$
    & $1.52_{\pm 0.09}$
    & $0.37_{\pm 0.11}$ \\
    \cmidrule{1-13}
    
    \textbf{Full UniReLo}
    & $\mathbf{97.8_{\pm 0.22}}$
    & $\mathbf{3.27_{\pm 0.26}}$
    & $\mathbf{0.15_{\pm 0.02}}$
    & $\mathbf{97.2_{\pm 0.45}}$
    & $\mathbf{2.74_{\pm 0.23}}$
    & $\mathbf{0.21_{\pm 0.05}}$
    & $\mathbf{93.2_{\pm 0.27}}$
    & $\mathbf{2.41_{\pm 0.15}}$
    & $\mathbf{0.26_{\pm 0.07}}$
    & $\mathbf{89.4_{\pm 0.27}}$
    & $\mathbf{1.86_{\pm 0.19}}$
    & $\mathbf{0.30_{\pm 0.10}}$ \\
    
    \midrule
    
    \multicolumn{13}{l}{\textbf{(3) Side Initial Postures}} \\
    \cmidrule{1-13}
    
    w/o TCG
    & $95.0_{\pm 1.73}$
    & $2.58_{\pm 0.38}$
    & $0.25_{\pm 0.09}$
    & $86.0_{\pm 1.22}$
    & $1.69_{\pm 0.23}$
    & $0.41_{\pm 0.05}$
    & $77.1_{\pm 1.19}$
    & $1.14_{\pm 0.21}$
    & $0.49_{\pm 0.09}$
    & $68.7_{\pm 1.48}$
    & $0.30_{\pm 0.09}$
    & $0.53_{\pm 0.10}$ \\
    \cmidrule{1-13}
    
    w/o TPP Init
    & $91.5_{\pm 1.27}$
    & $2.75_{\pm 0.32}$
    & $0.22_{\pm 0.07}$
    & $80.7_{\pm 0.97}$
    & $1.98_{\pm 0.19}$
    & $0.36_{\pm 0.02}$
    & $72.8_{\pm 1.86}$
    & $1.70_{\pm 0.38}$
    & $0.43_{\pm 0.04}$
    & $64.2_{\pm 1.99}$
    & $1.15_{\pm 0.19}$
    & $0.48_{\pm 0.06}$ \\
    \cmidrule{1-13}
    
    Single-scale AMP
    & $94.6_{\pm 1.34}$
    & $2.78_{\pm 0.27}$
    & $0.21_{\pm 0.05}$
    & $88.3_{\pm 1.15}$
    & $2.13_{\pm 0.40}$
    & $0.34_{\pm 0.03}$
    & $80.4_{\pm 1.92}$
    & $1.84_{\pm 0.37}$
    & $0.40_{\pm 0.03}$
    & $72.9_{\pm 1.43}$
    & $1.27_{\pm 0.31}$
    & $0.45_{\pm 0.09}$ \\
    \cmidrule{1-13}
    
    Hard-gated Routing
    & $95.8_{\pm 1.10}$
    & $2.84_{\pm 0.42}$
    & $0.20_{\pm 0.04}$
    & $89.5_{\pm 1.58}$
    & $2.15_{\pm 0.22}$
    & $0.32_{\pm 0.12}$
    & $81.5_{\pm 1.64}$
    & $1.89_{\pm 0.16}$
    & $0.38_{\pm 0.12}$
    & $75.3_{\pm 1.80}$
    & $1.32_{\pm 0.17}$
    & $0.41_{\pm 0.05}$ \\
    \cmidrule{1-13}
    
    \textbf{Full UniReLo}
    & $\mathbf{96.7_{\pm 0.42}}$
    & $\mathbf{3.06_{\pm 0.22}}$
    & $\mathbf{0.16_{\pm 0.06}}$
    & $\mathbf{95.8_{\pm 0.27}}$
    & $\mathbf{2.42_{\pm 0.25}}$
    & $\mathbf{0.27_{\pm 0.08}}$
    & $\mathbf{91.8_{\pm 0.45}}$
    & $\mathbf{2.12_{\pm 0.17}}$
    & $\mathbf{0.27_{\pm 0.02}}$
    & $\mathbf{89.2_{\pm 0.27}}$
    & $\mathbf{1.53_{\pm 0.14}}$
    & $\mathbf{0.36_{\pm 0.12}}$ \\
    
    \bottomrule
    \end{tabular}
    }
    
    \end{table*}

    \paragraph{Quantitative Ablation Results.}
    
    Table~\ref{tab:ablation_all} reports posture-resolved results under matched
    training and evaluation conditions. Full UniReLo achieves the best RTSR,
    ASM, and CTE under every evaluated terrain and initial-posture group.
    Averaged over prone, supine, and side postures, its RTSR reaches
    $98.0\%$, $97.1\%$, $93.3\%$, and $89.2\%$ on flat ground, gravel,
    the $10^\circ$ slope, and the $15^\circ$ slope, respectively.
    
    Removing terrain-pose plasticity-aware initialization causes the largest
    RTSR reduction. The posture-averaged reduction increases from $5.60$
    percentage points on flat ground to $24.03$ percentage points on the
    $15^\circ$ slope, indicating that terrain-sensitive fallen-state sampling
    becomes increasingly important as recovery difficulty grows.
    
    Removing terrain-conditioned recovery guidance mainly degrades ASM and
    subsequent command tracking. Compared with the full model, the
    posture-averaged ASM decreases by $0.52$\,cm, $0.79$\,cm, $1.02$\,cm,
    and $1.24$\,cm across the four terrain conditions. The same variant also
    produces the largest overall CTE, showing that unsuitable support
    configurations directly affect the resumption of commanded locomotion.
    
    Single-scale AMP and Hard-gated Routing both reduce RTSR and increase CTE.
    The former removes temporal specialization, whereas the latter removes
    continuous coordination among the three motion priors. Together with Fig.~\ref{fig:gate_comparison}, these results support both
multi-scale temporal supervision and continuous progress-dependent gating.
    
    \subsection{Real-World Evaluation}
    \label{subsec:realworld}
    
    The simulation-trained UniReLo policy is deployed on a physical 29-DoF
    Unitree G1 humanoid. The evaluation includes indoor flat ground, outdoor
    gravel, outdoor grass, and outdoor slopes of $10^\circ$ and $15^\circ$.
    Only outdoor grass is absent from the simulation training terrains. All
    full-model experiments use the same policy checkpoint and deployment
    parameters without terrain-specific retuning or online terrain labels.
    
    A sequence is considered successful only if the robot recovers from the
    fallen state, resumes sustained command-following locomotion, and avoids a
    secondary fall within the evaluation horizon.
    
    \paragraph{Real-World Failure Analysis.}
    
    \begin{figure*}[!t]
    \centering
    
    \begin{subfigure}{0.85\textwidth}
    \centering
    \includegraphics[width=\linewidth]{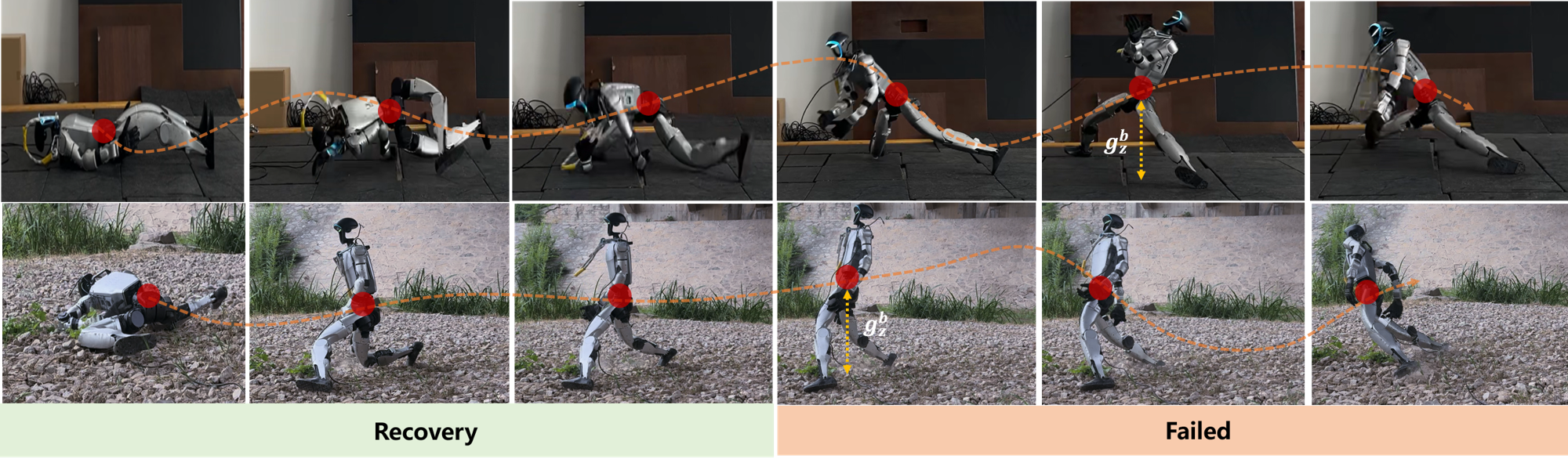}
    \caption{Hard-gated Routing.}
    \label{fig:extreme_slope:a}
    \end{subfigure}
    
    \vspace{0.3em}
    
    \begin{subfigure}{0.85\textwidth}
    \centering
    \includegraphics[width=\linewidth]{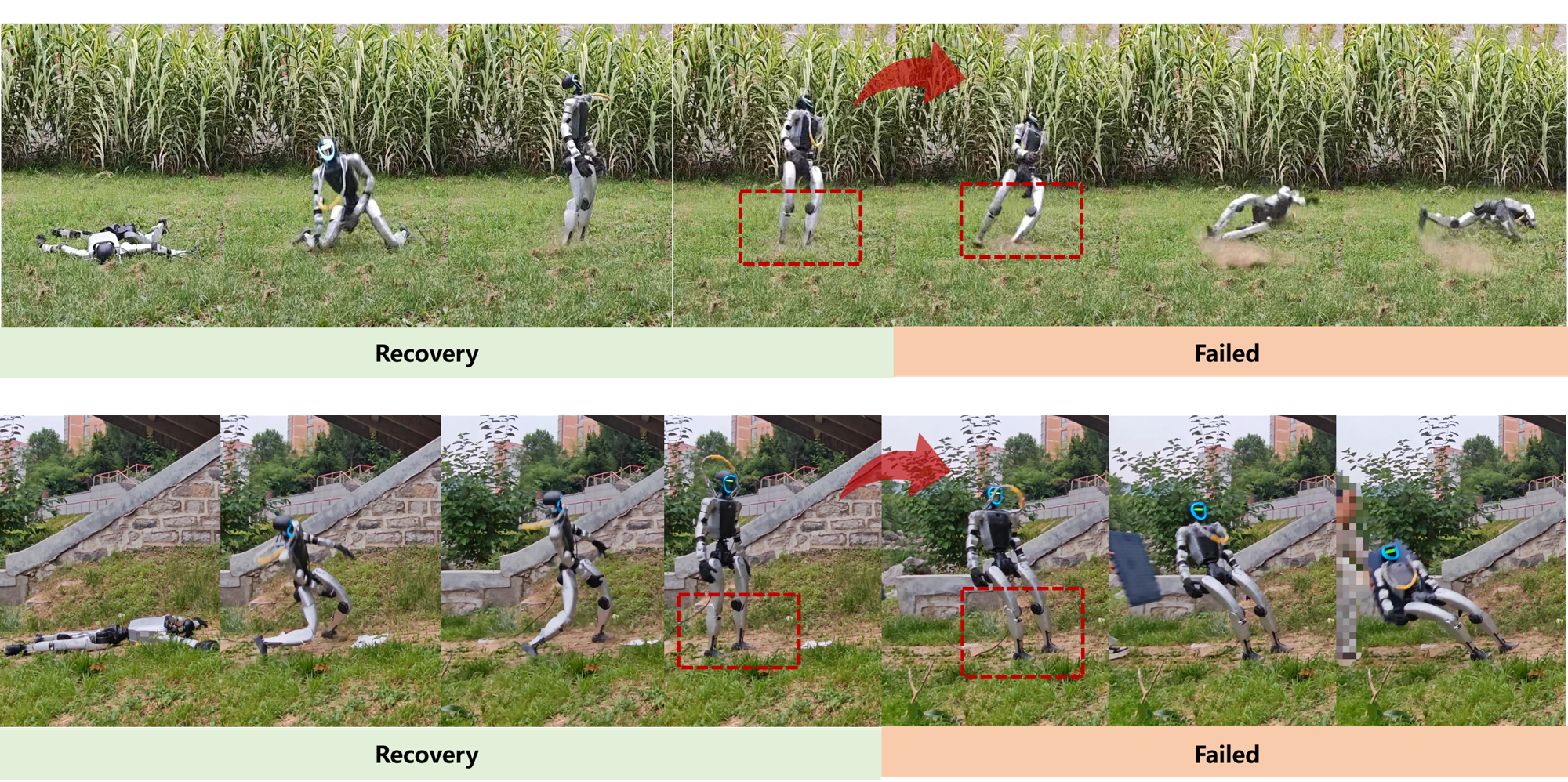}
    \caption{w/o TCG.}
    \label{fig:extreme_slope:b}
    \end{subfigure}
    
    \caption{
    Representative real-world failures after removing key components of
    UniReLo. (a) Hard-gated Routing produces abrupt behavior near the routing
    boundary, resulting in discontinuous center-of-mass motion, repeated
    recovery actions, and secondary falls. Red dots and dashed orange curves
    indicate approximate center-of-mass positions and trajectories.
    (b) Removing terrain-conditioned recovery guidance produces insufficient contact and
    posture adaptation on grass and inclined terrain. Red dashed boxes highlight
    unstable foot configurations immediately before loss of balance.
    }
    \label{fig:extreme_slope}
    \end{figure*}
    
    Fig.~\ref{fig:extreme_slope:a} shows that discrete routing is sensitive to
    fluctuations near its predefined transition boundary. Abrupt changes in the
    dominant motion prior repeatedly drive the robot toward recovery-like actions
    after it has temporarily become upright. The resulting center-of-mass
    oscillation eventually causes a secondary fall.
    
    Fig.~\ref{fig:extreme_slope:b} shows a different failure mode after
    removing terrain-conditioned recovery guidance. The robot may reach the target body
    height, but poorly positioned feet and insufficient terrain-dependent postural adaptation prevent the robot from establishing a terrain-compatible support state for continued locomotion. These cases provide hardware evidence that uprightness alone does not guarantee successful recovery in field environments.
    
    \paragraph{Field Robustness and Recovery-to-Locomotion.}
    
    We first evaluate UniReLo under external perturbations during locomotion.
    As shown in Fig.~\ref{fig:teaser}(b), (d), and (f), the robot is tested on
    outdoor grass, outdoor gravel, and a $15^\circ$ outdoor slope. Under
    moderate disturbances, it maintains locomotion through coordinated
    whole-body regulation and adaptive foot placement. Under fall-inducing
    disturbances, it performs whole-body recovery, re-establishes
    terrain-compatible support, and resumes commanded walking. The grass experiment evaluates transfer to an unseen deformable surface, whereas gravel and inclined terrain evaluate robustness under irregular
    contact and terrain inclination.
    
   \begin{figure*}[!t]
\centering
\includegraphics[width=0.90\textwidth]{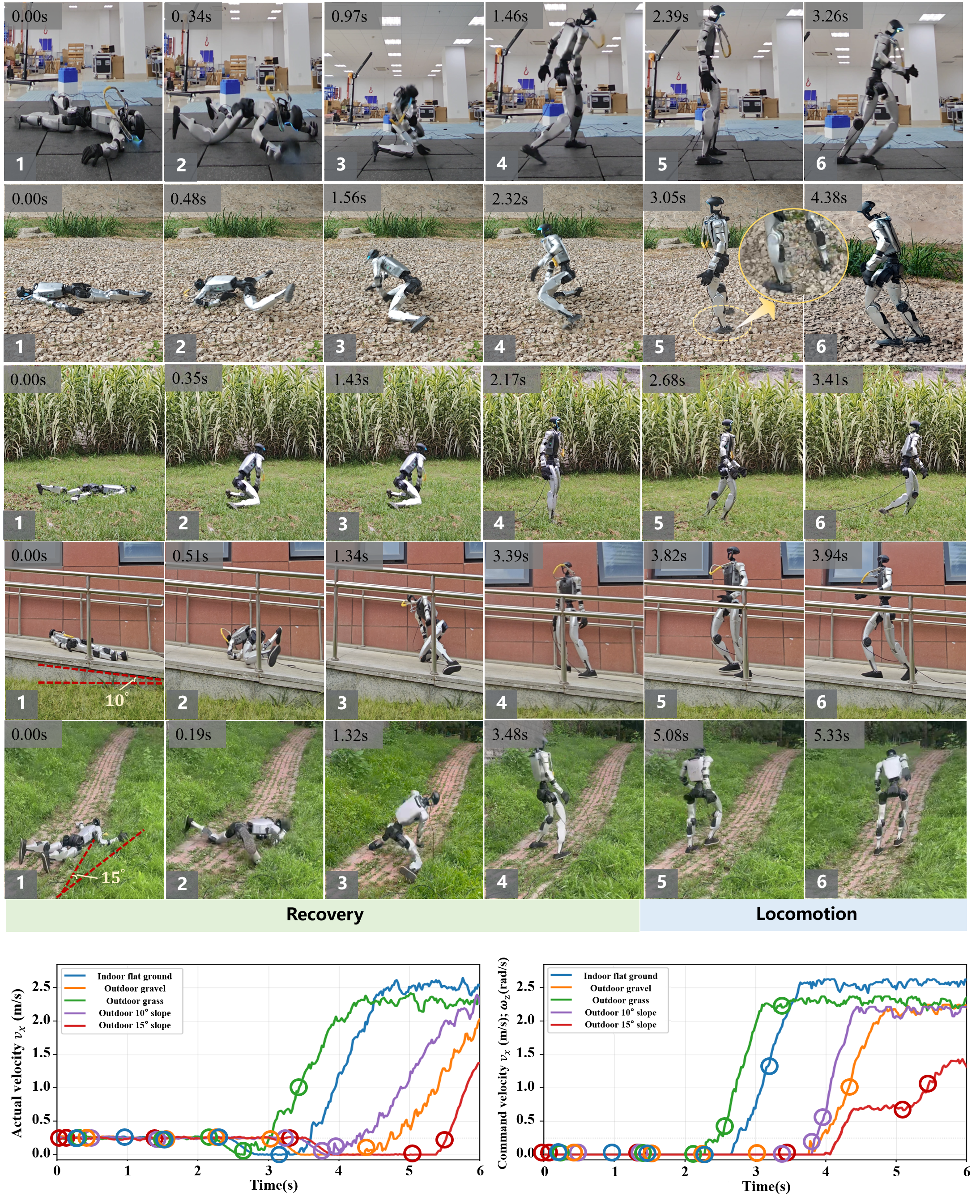}
\caption{
Real-world recovery-to-locomotion behaviors and corresponding velocity
profiles across five terrain conditions. From top to bottom, the image
sequences show recovery and subsequent locomotion on indoor flat ground,
outdoor gravel, outdoor grass, and outdoor slopes of $10^\circ$ and
$15^\circ$. Timestamps indicate elapsed time after policy activation.
The two plots at the bottom show the measured forward velocity $v_x$ and
the remote forward-velocity command $v_x^{\mathrm{cmd}}$, respectively.
Colored curves denote the five terrain conditions, and circular markers
associate the velocity trajectories with selected frames in the image
sequences.
}
\label{fig:real_world_combined}
\end{figure*}

Fig.~\ref{fig:real_world_combined} jointly presents the qualitative recovery sequences and the corresponding forward-velocity profiles across all five real-world conditions. The image sequences reveal terrain-dependent recovery strategies. Flat ground permits a relatively direct rise into locomotion, whereas gravel requires additional foot-placement adjustments under irregular contact. On grass, the robot adopts more compliant whole-body motion to accommodate the deformable surface. On inclined terrain, UniReLo establishes terrain-compatible support through increased knee flexion, terrain-aligned posture, and a longer stabilization interval. These adaptations become more pronounced as the inclination increases from $10^\circ$ to $15^\circ$.

The velocity plots at the bottom further quantify the transition from recovery to command-following locomotion. The colored circular markers associate the trajectories with selected recovery frames shown above. Indoor flat ground and outdoor grass exhibit earlier forward progression, whereas gravel and the inclined terrains require longer support reorganization before sustained locomotion is established. On the $15^\circ$ slope, the robot briefly performs an in-place heading correction after becoming upright, which delays the increase in measured forward velocity despite the continued forward command. Forward locomotion then resumes once the heading is adjusted.

\section{CONCLUSIONS}
\label{sec:conclusions}

We presented UniReLo, a unified proprioceptive policy for humanoid fall
recovery and subsequent locomotion across diverse terrains. Continuously gated multi-scale motion priors coordinate frame-, sequence-, and gait-level supervision according to recovery progress, enabling a smooth transition from whole-body recovery to velocity-commanded locomotion. Terrain-conditioned recovery guidance further encourages the robot to establish terrain-compatible support states capable of sustaining continued motion, rather than relying on upright posture alone. Simulation and real-world evaluations demonstrate reliable recovery-to-locomotion transitions, disturbance robustness, and cross-terrain transfer across flat ground, gravel, grass, and inclined terrains without terrain-specific controller switching or parameter retuning.

Future work will incorporate exteroceptive and tactile sensing and expand the diversity of terrains and motion priors to improve adaptation to deformable and out-of-distribution surfaces.



\bibliographystyle{IEEEtran}
\bibliography{reference}

\end{document}